\documentclass[journal]{IEEEtran}

\usepackage{amsmath}
\usepackage{graphicx}
\usepackage{subfig} 
\usepackage{tabu}  

\usepackage[colorlinks,linkcolor=blue,anchorcolor=blue,citecolor=green]{hyperref}
\usepackage{multirow}
\usepackage{dsfont}
\usepackage{stfloats}
\usepackage{amsmath}
\usepackage{cite}
\usepackage{booktabs}
\usepackage{algorithm}
\usepackage{algorithmic}
\usepackage{amssymb}
\usepackage{makecell}
\usepackage{bm}
\usepackage{caption}
\usepackage{xcolor}
\usepackage{tikz}
\definecolor{mypink}{RGB}{243, 196, 255} 
\definecolor{myblue}{RGB}{207, 233, 244} 
\definecolor{tp}{RGB}{0,118,121} 
\definecolor{tn}{RGB}{247,238,238} 
\definecolor{fp}{RGB}{255,171,96} 
\definecolor{fn}{RGB}{239,95,112} 
\definecolor{bareland_color}{rgb}{0.502,0.0,0.0}
\definecolor{cropland_color}{rgb}{0.655, 0.733,0.106}
\definecolor{vegetation_color}{rgb}{0.274,0.710,0.474}
\definecolor{water_color}{rgb}{0.110,0.549,0.741}
\definecolor{building_color}{rgb}{0.710, 0.274, 0.274}
\definecolor{road_color}{rgb}{0.870,0.721,0.274}
\definecolor{developed_space_color}{rgb}{0.580,0.580,0.580}
\definecolor{background_color}{rgb}{0.949,0.937,0.914}

\ifCLASSINFOpdf

\else
 
\fi

\begin{document}

\captionsetup{font = {small}}

\title{ObjFormer: Learning Land-Cover Changes From Paired OSM Data and Optical High-Resolution Imagery via Object-Guided Transformer}

\author{
        Hongruixuan~Chen,~\IEEEmembership{Student Member,~IEEE,}
        Cuiling~Lan,~\IEEEmembership{Member,~IEEE,}
        Jian~Song,
        Clifford Broni-Bediako,~\IEEEmembership{Member,~IEEE,}
        Junshi~Xia,~\IEEEmembership{Senior Member,~IEEE,}
        Naoto~Yokoya,~\IEEEmembership{Member,~IEEE}

\thanks{This work was supported in part by the Council for Science, Technology and Innovation (CSTI), the Cross-ministerial Strategic Innovation Promotion Program (SIP), Development of a Resilient Smart Network System against Natural Disasters (Funding agency: NIED), the JSPS, KAKENHI under Grant Number 22H03609, JST, FOREST under Grant Number JPMJFR206S, Microsoft Research Asia, Next Generation AI Research Center, The University of Tokyo, and the Graduate School of Frontier Sciences, The University of Tokyo through the Challenging New Area Doctoral Research Grant (Project No. C2303). \emph{(Corresponding author: Naoto Yokoya)}}
\thanks{Hongruixuan Chen, Jian Song, and Naoto Yokoya are with Graduate School of Frontier Sciences, The University of Tokyo, Chiba, Japan (e-mail: qschrx@gmail.com; song@ms.k.u-tokyo.ac.jp; yokoya@k.u-tokyo.ac.jp).}
\thanks{Cuiling Lan is with Microsoft Research Asia, Beijing, China (e-mail: culan@microsoft.com).}
\thanks{Clifford Broni-Bediako and Junshi Xia and RIKEN Center for Advanced Intelligence Project (AIP), RIKEN, Tokyo, 103-0027, Japan (e-mail: akosabroni@gmail.com; junshi.xia@riken.jp).}

}

\markboth{TRANSACTIONS ON GEOSCIENCE AND REMOTE SENSING, 2024}%
{Shell \MakeLowercase{\textit{\emph{et al.}}}: Bare Demo of IEEEtran.cls for IEEE Journals}

\maketitle

\begin{abstract}
Optical high-resolution imagery and OpenStreetMap (OSM) data are two important data sources of land-cover change detection (CD). Previous related studies focus on utilizing the information in OSM data to aid the CD on optical high-resolution images. This paper pioneers the direct detection of land-cover changes utilizing paired OSM data and optical imagery, thereby expanding the scope of CD tasks. To this end, we propose an object-guided Transformer (ObjFormer) by naturally combining the object-based image analysis (OBIA) technique with the advanced vision Transformer architecture. This combination can significantly reduce the computational overhead in the self-attention module without adding extra parameters or layers. Specifically, ObjFormer has a hierarchical pseudo-siamese encoder consisting of object-guided self-attention modules that extracts multi-level heterogeneous features from OSM data and optical images; a decoder consisting of object-guided cross-attention modules can recover land-cover changes from the extracted heterogeneous features. Beyond basic binary change detection (BCD), this paper raises a new semi-supervised semantic change detection (SCD) task that does not require any manually annotated land-cover labels to train semantic change detectors. Two lightweight semantic decoders are added to ObjFormer to accomplish this task efficiently. A converse cross-entropy (CCE) loss is designed to fully utilize negative samples, contributing to the great performance improvement in this task. A large-scale benchmark dataset called OpenMapCD containing 1,287 map-image pairs covering 40 regions on six continents is constructed to conduct detailed experiments. The results show the effectiveness of our methods in this new kind of CD task. Additionally, case studies in two Japanese cities demonstrate the framework's generalizability and practical potential. The OpenMapCD and source code are available in \url{https://github.com/ChenHongruixuan/ObjFormer}.
\end{abstract}


\IEEEpeerreviewmaketitle

\section{Introduction}\label{sec:1}
\par \IEEEPARstart{T}{he} Earth's surface undergoes continuous changes, influenced by a myriad of natural and anthropogenic factors. Leveraging remote sensing technology for multi-temporal Earth observation, change detection (CD) technology has emerged as a pivotal tool, enabling the macroscopic monitoring and interpretation of changes in the Earth's surface through the use of airborne and spaceborne sensors \cite{Singh1989}. At present, CD is indispensable in various fields, including land-cover and land-use analysis, urban expansion studies, ecological environment monitoring, and disaster management \cite{Coppin2004, Xian2009, ZHENG2021Building, Li2024China}.

\par Optical remote sensing imagery has been one of the most important data sources for CD. The advent of sensors capable of capturing optical high-resolution images, such as SPOT, WorldView, and QuickBird, marked a significant advancement, offering unprecedented clarity in ground information, such as edge and texture details. This enhancement paved the way for finer change analysis in practical applications, notably in areas like building footprint updating, deforestation monitoring, precision agriculture, and building damage assessment \cite{Finer2018Combating, guo2021deep, ZHENG2021Building, xiao2023degrade}. However, the high spatial resolution introduces new challenges, such as increased reflectance variance within the class, georeferencing accuracy, and different acquisition characteristics \cite{Hussain2013}, posing hurdles for traditional pixel-based methods, which only focus on the meticulous extraction and utilization of each pixel's spectral information \cite{Sharma2007, Nielsen1997, Wu2017b, Lei2023Simple, Zhu2014Continuous}. Addressing these challenges calls for the integration of advanced analytical tools and techniques that delve beyond spectral information to encompass spatial textural details present in optical high-resolution images. A prominent solution is the object-based image analysis (OBIA) method \cite{Hussain2013}, which operates on homogeneous regions, called objects, generated through unsupervised image segmentation algorithms. This approach groups numerous pixels into objects, forming the basis for subsequent change analyses, effectively mitigating the “salt-and-pepper” noise issue encountered in pixel-based methods \cite{Comber2004Application, Liu2021, BONTEMPS2008object, Chen2014Assessment, Zhang2017a, Sun2021a}. These methods have gained traction, establishing themselves as a prevalent framework for CD in optical high-resolution images \cite{Hussain2013}. 

\par In recent years, deep learning has revolutionized the field of computer vision, extending its reach to encompass various remote sensing tasks, including CD in optical high-resolution images. A plethora of studies have explored the potential of diverse deep learning architectures, including convolutional neural networks (CNNs) \cite{Zhan2017}, recurrent neural networks (RNNs) \cite{Lyu2016}, graph convolutional networks (GCNs) \cite{Wu2021multiscale,chen2022unsupervised}, generative adversarial networks (GANs) \cite{Wu2023Fully}, and hybrid methods that integrate multiple architectures \cite{Chen2019a, Tang2022}. Notably, fully convolutional network (FCN)-based approaches have become mainstream, demonstrating impressive results in large-scale benchmark datasets and practical applications \cite{CayeDaudt2018, Zhang2020, Fang2022SNUNet, guo2021deep, Zheng2022, xiao2022msdtgp, Wu2023Fully, Han2023Change, CAO2023full, Chen2023Exchange}. Some studies tried to combine the OBIA techniques with deep learning architectures, further enhancing CD and damage assessment tasks \cite{Liu2021, ZHENG2021Building, Zhang2023ESCNet}. More recently, the vision Transformer architecture \cite{dosovitskiy2020image} has been introduced to this field, whose ability to capture non-local pixel relationships is well suited for extracting spatial contextual information from optical high-resolution images \cite{Chen2022Remote, Zhang2022SwinSUNet, Bandara2022Transformer, LiTransUNetCD2022, Zhang2023Relation}. In addition to using the above deep learning architectures as feature extractors, many studies \cite{Zheng2022, Han2023Change, Ding2022Temporal, Ding2024Joint, Huang2024Spatiotemporal} explored how to appropriately model the spatio-temporal relationships of multi-temporal images to achieve better change detection results.

\par Beyond remote sensing imagery acquired by airborne and spaceborne sensors, geographic information data are often used for CD \cite{Lu2004}. A prime example of this is OpenStreetMap (OSM). As a collaborative platform, OSM allows users worldwide to input and update geographical and infrastructure data, offering a rich and continuously updated source of information that can be leveraged for CD tasks. Integrating OSM data with optical high-resolution imagery not only mitigates some of the challenges associated with optical imagery but also brings forth new avenues for monitoring and understanding changes occurring on the Earth's surface \cite{OpenStreetMap2021Vargas}. Most current research involved in OSM data used it as a supplementary information source, such as land-cover/land-use information and geometric information, to assist the CD task \cite{Rienow2022Detecting, Shi2020Land, Wang2022Graph}. Some recent studies \cite{albrecht2020change, Liao2023BCE} tried to update building footprints by performing the CD between OSM data or building labels derived from historical maps and up-to-date optical imagery. However, these studies have only addressed building CD rather than more general land-cover CD. In addition, they have been more limited in geographic scope.

\par To the best of our knowledge, no research positions OSM data as a primary data source and directly performs binary and detailed semantic land-cover change analysis on paired optical imagery and OSM data. Exploring such a topic is undoubtedly of practical significance to both the remote sensing and geographic information communities. On the one hand, the development of appropriate techniques will allow the community to directly utilize a richer set of historical and future data from multiple sources to analyze finer land-cover changes at a higher temporal resolution. In addition, OSM provides a large number of free labels with detailed semantic information \cite{SCHULTZ2017Open, Shi2020Land}, allowing us to perform further SCD at a lower labeling cost instead of being limited to basic BCD. On the other hand, the ability to analyze changes using OSM data and the most recent optical images provides an automatic mechanism for updating and correcting map data. This implication extends to research into automated systems that can integrate CD results into existing map databases, ensuring that geographical information systems are consistently up-to-date with lower manual labor. Given the above, this study attempts to bridge the research gap.

\par However, performing land-cover CD on paired optical remote sensing imagery and OSM data poses inherent challenges, primarily due to their differing properties. Traditional CD tasks involve analyzing a pre-event and a post-event image, both bearing a clear temporal order and physical significance, i.e., a record of certain physical properties of the Earth's surface before and after a occurred change event. However, in our task, the inputs to the detector will be OSM data and optical high-resolution imagery, with no clear temporal order between them. Optical images are raster-based, capturing continuous reflectance values across specific wavelengths, whereas OSM data provide vector-based symbolic representations of real-world features, such as roads and buildings. Such a large domain gap between these two data forms presents a formidable challenge. Also, the OSM data itself carries certain virtual attributes, which do not necessarily truly reflect the surface coverage due to the mechanism of crowdsourcing updates. In order to solve the above problems, designing a powerful architecture is necessary. Currently, vision Transformer has achieved very promising results in the field of CD \cite{Chen2022Remote, Zhang2022SwinSUNet}, with great potential to be applied to our task. However, there are some problems with vision Transformer architecture. The core of the Transformer lies in its self-attention mechanism\footnote{Strictly speaking, the Transformer employs multi-head self-attention. Unless otherwise noted, all terms regarding self-attention in this paper refer to multi-head self-attention.}, which can effectively model non-local relationships between pixels. This part tends to generate considerable computational overhead and GPU memory burden \cite{dosovitskiy2020image, Rao2021DynamicViT, Wang2021Pyramid}. This shortcoming will be magnified in visually dense prediction tasks, as it is required to maintain feature maps with relatively higher resolution \cite{Wang2021Pyramid}. 

\par In this paper, we simultaneously focus on supervised BCD and semi-supervised SCD using paired OSM data and optical high-resolution imagery as the data sources. The first sizeable open-source benchmark dataset called OpenMapCD was constructed for these two tasks to facilitate this study and subsequent related research. We propose a novel Transformer framework called ObjFormer, where we combine the OBIA method with the self-attention mechanism, without introducing any new parameters. In this way, the computational cost and GPU memory burden of the Transformer architecture are significantly reduced, and low-level information is naturally introduced to guide the non-local relationship modeling in deep layers of the network. Furthermore, for the semi-supervised SCD task raised in this study, a converse cross-entropy (CCE) loss is designed to fully utilize the information in negative samples (changed areas), \emph{i.e.}, although we do not know what the area is, we know what the area is not.

\begin{table}[!t]
  \renewcommand{\arraystretch}{1.3}
\caption{\centering{List of key abbreviations.}}\label{tbl:abbre} 
  \centering
  \begin{tabular}{c c }
  \toprule
    \hline	
   \textbf{Abbreviation}	& \textbf{Meaning}  \\
    \hline\hline
  CD &	Change detection	 \\
 BCD & Binary change detection  \\
  SCD & Semantic change detection  \\
OBIA & Object-based image analysis \\
OSM & OpenStreetMap \\ 
ObjFormer & Object-guided Transformer\\ 
CCE  &Converse cross-entropy \\ 
GSD & Ground sampling distance \\ 
SLIC & Simple linear iterative clustering \\
MACs &  Multiply-accumulate operations \\ 
    \hline
    \bottomrule
  \end{tabular}
\end{table}

\par In summary, the main contributions of this paper are as follows:
\begin{enumerate}
    \item We propose a new form of CD, \emph{i.e.}, detecting general land-cover changes directly from paired optical imagery and OSM data, which includes a basic supervised BCD task as well as a further semi-supervised SCD task.
    \item For this new CD form, we present an object-guided Transformer architecture that can accurately detect land-cover changes from these two heterogeneous data forms. Our architecture naturally integrates the self-attention mechanism with OBIA techniques, significantly reducing computational overhead.
    \item A converse cross-entropy loss is devised for the semi-supervised SCD task, which can effectively mine the information of negative samples and thus significantly improve the performance of different methods.
    \item The first multimodal benchmark dataset called OpenMapCD covering 40 regions on six continents is constructed for detailed experiments. It will be open-sourced to facilitate subsequent research in the community.
\end{enumerate}

\par The remainder of this paper is organized as follows. Section \ref{sec:2} presents our OpenMapCD dataset constructed for this study together with a detailed presentation of the two selected study sites for testing. Section \ref{sec:3} elaborates on the proposed ObjFormer architecture. The experimental design and results are provided in Sections \ref{sec:4} and \ref{sec:5}, respectively. Section \ref{sec:6} discusses the limitations of the current study and future work. Section \ref{sec:7} draws the conclusion. In addition, Table \ref{tbl:abbre} summarizes the meanings of key abbreviations used in this paper for ease of retrieval.

\section{Study area and data}\label{sec:2}

\subsection{Large-scale Benchmark Dataset}
\par For the paired OSM data and optical imagery-based CD tasks, we built a large-scale map-optical imagery pair CD dataset called OpenMapCD. This facilitates our study and is expected to inspire more future investigation. 

\begin{figure}[!t]
  \centering
\includegraphics[width=3.55in]{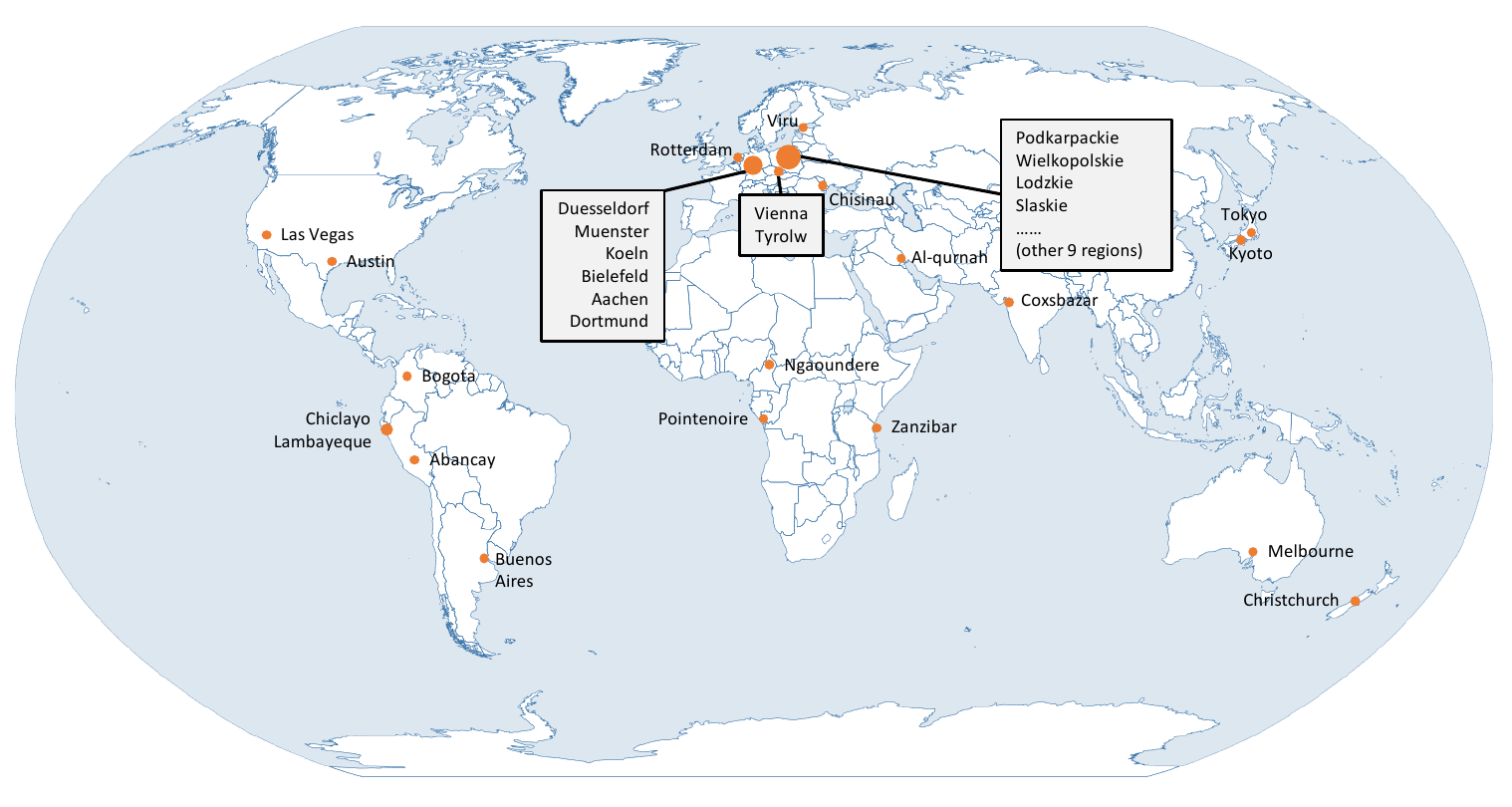}
  \caption{Data distribution regions for the OpenMapCD dataset, covering 40 regions across six continents. }
  \label{data_distribution}
\end{figure}

\begin{table}[!t]
  \renewcommand{\arraystretch}{1.3}
  \caption{\centering{Correspondence between OSM categories and specific land cover categories.}}
  \label{tbl:label_mapping}
  \centering
  \begin{tabular}{l | p{5.4cm}}
  \toprule
    \hline	
   \textbf{Land-cover category} &  \textbf{Corresponding OSM feature} \\
    \hline\hline
   \tikz[baseline=-0.09cm]{\fill[bareland_color] (-0.2,-0.2) rectangle (0.15cm,0.15cm);} 
bareland  &	quarry, landfill, brownfield, fell, sand, scree, beach, mud, glacier, rock, cliff \\
\tikz[baseline=-0.09cm]{\fill[cropland_color] (-0.2,-0.2) rectangle (0.15cm,0.15cm);} cropland &	farmland, farm, farmyard, greenhouse, vineyard, orchard \\
\tikz[baseline=-0.09cm]{\fill[vegetation_color] (-0.2,-0.2) rectangle (0.15cm,0.15cm);} vegetation & forests, woods, grass, greenfield, scrub, heath, grassland, meadow, golf course
 \\
\tikz[baseline=-0.09cm]{\fill[water_color] (-0.2,-0.2) rectangle (0.15cm,0.15cm);} water &	river, creek, lakes, reservoirs, canal, pond \\
\tikz[baseline=-0.09cm]{\fill[road_color] (-0.2,-0.2) rectangle (0.15cm,0.15cm);} road &	various levels of roads, railways and bridges  \\
\tikz[baseline=-0.09cm]{\fill[building_color] (-0.2,-0.2) rectangle (0.15cm,0.15cm);} building &	building, other special buildings (e.g., hospitals, pharmacies, etc.) \\
\tikz[baseline=-0.09cm]{\fill[developed_space_color] (-0.2,-0.2) rectangle (0.15cm,0.15cm);} developed spaces & playground, parking lot, square, retail area, industrial area, commercial area \\
\tikz[baseline=-0.09cm]{\fill[background_color] (-0.18,-0.18) rectangle (0.15cm,0.15cm);} background & basemap and others  \\

    \hline
    \bottomrule
  \end{tabular}
\end{table}
\par The optical high-resolution images in the OpenMapCD dataset are sourced from the OpenEarthMap dataset\footnote{https://open-earth-map.org/} \cite{Xia2023OpenEarthMap}, an open-sourced large-scale land-cover mapping dataset. This dataset encompasses 5,000 optical high-resolution images with the size of 1,024$\times$1,024 pixels and 0.25-0.5m ground sampling distance (GSD), containing three bands of RGB. According to the geographic coordinate and projected coordinate systems of these optical images, we downloaded the corresponding OSM data\footnote{https://www.openstreetmap.org/} of these areas covered by optical images with the zoom level of 18 and rasterized them, ensuring resolution consistency with the optical images. Given the disparate editorial situations across different regions in OSM data, with some areas meticulously mapped while others remain scarcely edited, we opted to exclude map-image pairs where over 80$\%$ areas consisted of base maps, i.e., not edited, thereby guaranteeing a sufficient representation of both changed and unchanged labels in the samples. The finalized dataset comprises 1,287 map-image pairs, each with the size of 1,024$\times$1,024 pixels (for most of them), spanning 40 regions across six continents, as illustrated in Fig. \ref{data_distribution}, ensuring the geographic diversity of the dataset. For the specific land-cover types, the dataset encompasses seven prevalent land-cover categories: bareland, cropland, vegetation, water, road, building, and other developed spaces, categories frequently associated with natural and man-made change events \cite{Yang2022Asymmetric, Xia2023OpenEarthMap}. Compared to the above seven land-cover categories, OSM data contains more detailed information. For example, for roads that land-cover category, OSM data further defines a variety of different classes and levels of roads. In order to obtain land-cover change labels, we need to establish a mapping relationship between OSM features and specific land-cover categories. We draw on the information summarized in \cite{SCHULTZ2017Open}, while incorporating the land-cover categories involved in our study. Table \ref{tbl:label_mapping} summarizes the correspondence between land-cover categories and OSM features. Finally, we perform an XOR operation on OSM data and land-cover labels of optical images to generate high-quality pixel-wise CD labels, delineating changed, unchanged, and background areas. 

\begin{figure*}[!t]
  \centering
\includegraphics[width=6.2in]{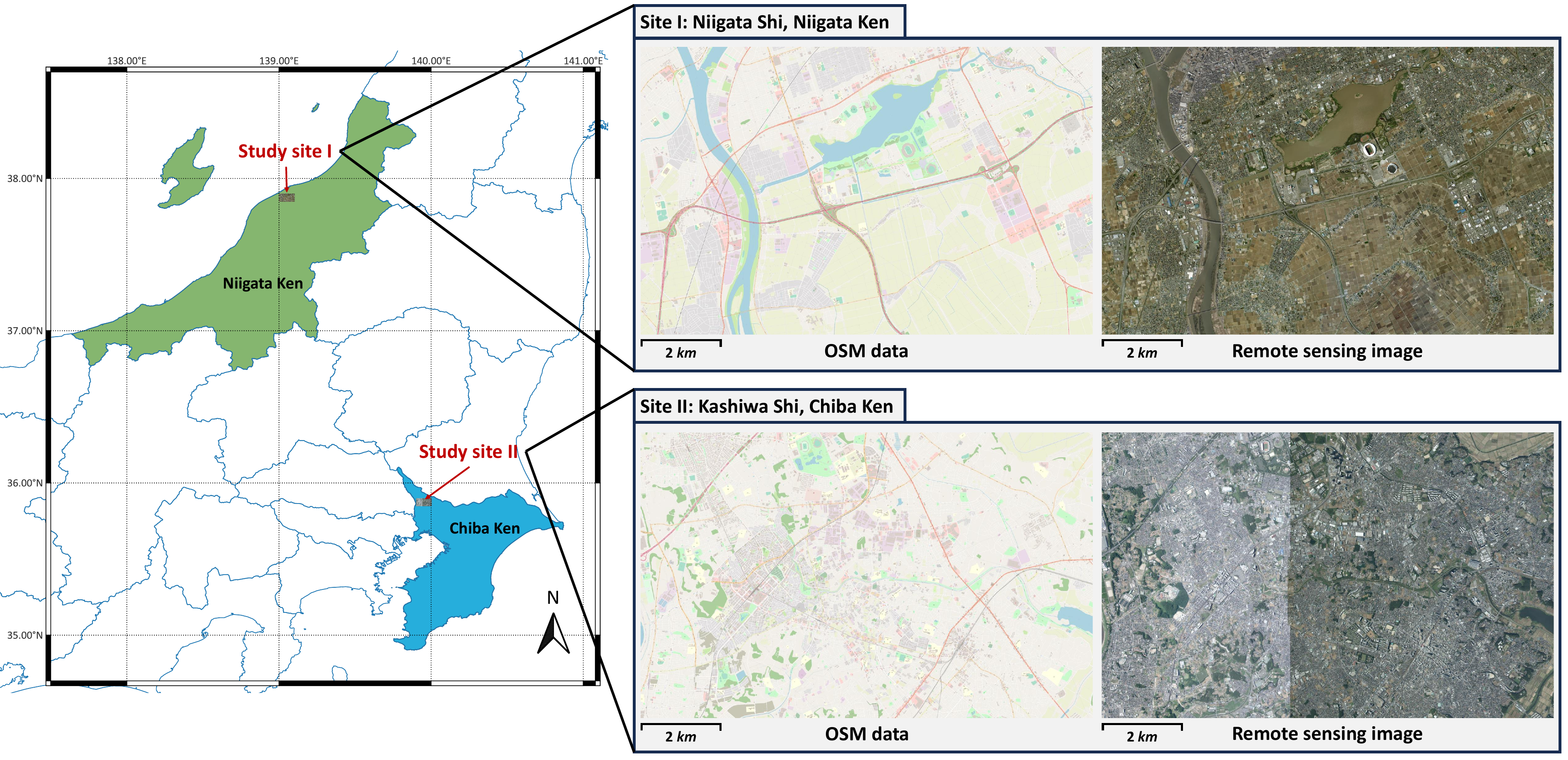}
  \caption{Locations of two study sites and corresponding OSM data and optical high-resolution imagery for each of them. Note that the two study sites are not covered by the OpenMapCD dataset.}
  \label{study_site}
\end{figure*}

\par For training and evaluating detectors, we partitioned the OpenMapCD dataset into a training set and a test set containing 901 (70$\%$) and 386 (30$\%$) pairs, respectively. In the subsequent experiments, the training set will be used to train different change detectors, and the test set will be used for accuracy assessment, benchmark comparison, ablation studies, hyperparameter discussions, etc. The OpenMapCD dataset will be open-sourced as a publicly available benchmark to facilitate follow-up research.

\subsection{Local Study Sites}\label{sec:local_study_site}

\par To validate the generalizability of the models trained using our OpenMapCD dataset and to illustrate their potential real-world applications, we identified additional study sites not encompassed in the OpenMapCD dataset to conduct experiments. Given the observed increase in OSM participants in Japan, yet with confined areas receiving updates\footnote{https://qiita.com/kouki-T/items/9b0e72710f3e8ca3dc4c}, Japan presents favorable conditions for selecting study sites with substantial differences between OSM data and optical imagery. 

\par As shown in Fig. \ref{study_site}, the study site I is in Niigata City, located in the northeastern part of Niigata Prefecture, Japan, and is the capital of Niigata Prefecture. It is the most populous city on the west coast of Honshu and the second most populous city in the Chūbu region after Nagoya. As a port city, Niigata is a key point of land and water transportation in Japan. The study site II is Kashiwa City, located in the northwest part of Chiba Prefecture, Japan. Kashiwa city is an important commercial city in the eastern part of the capital metropolitan area. 

\par For the test data, optical high-resolution images were supplied by The Geospatial Information Authority of Japan (GSI)\footnote{https://www.gsi.go.jp/ENGLISH/index.html/}, with site I covering an 81.28 $km^{2}$ region with 18,944$\times$12,036 pixels, and site II covering a 78.53 $km^{2}$ region with 18,688$\times$11,776 pixels. GSD of both optical images is 0.6m/pixel. Leveraging the geographic coordinates of these images, we procured the corresponding OSM data, maintaining resolution consistency with the optical images during the rasterization process. From the optical images, we can see that site I includes part of the urban area and part of the rural area, while site II is mainly the inner-city area. The two sites demonstrate significantly different land-cover distributions. This is also more conducive for us to verify the generalization of the method. Furthermore, the OSM data for both regions have not been adequately updated. There are many regions that have not been edited or those that have not been updated promptly. We will use these two large-scale map-image pairs to demonstrate practical applications of the methodology, including land-cover mapping, semantic change analysis, and map updating. Similarly, test data and their labels on both study sites will be open-sourced.

\section{Methodology}\label{sec:3}
\subsection{Problem Statement}\label{sec:3.1}
\par Based on the outputs of change detectors, the CD tasks can be categorized into BCD and SCD. The output of the former is a subset of the output of the latter, but the latter requires more annotation information. In our study, the constructed OpenMapCD dataset can support us to perform SCD directly. However, given the breadth of the community for BCD task, and the necessity for this work to provide benchmarks for subsequent work, we concentrate on both BCD and SCD tasks. In the following, we describe the formal definition and objective of each task in detail.

\subsubsection{\textbf{Supervised binary change detection}}
\par BCD is the basic task in the field of CD, aiming to provide binary maps of “change / non-change” to reflect information on changes in land-cover objects. In the context of this paper, the BCD task can be formally stated as follows: given a training set represented as $\mathcal{D}^{bcd}_{train} = \left\{\left(\mathbf{X}^{osm}_{i}, \mathbf{X}^{opt}_{i}, \mathbf{Y}^{bcd}_{i}\right)\right\}_{i=1}^{N_{train}}$, where $\mathbf{X}^{osm}_{i}\in \mathbb{R}^{\mathrm{H}\times \mathrm{W}\times 3}$ is the $i$-th rasterized OSM data, $\mathbf{X}^{opt}_{i}\in \mathbb{R}^{\mathrm{H}\times \mathrm{W}\times \mathrm{C}^{opt}}$ is the corresponding optical high-resolution image, and $\mathbf{Y}^{bcd}_{i} \in \left\{0,1\right\}^{\mathrm{H}\times \mathrm{W}}$ is the associated change reference map, the goal is to train a detector $\mathcal{T}^{bcd}$ capable of precisely identifying land-cover changes from paired OSM data and optical imagery in new sets.

\subsubsection{\textbf{Semi-supervised semantic change detection}}\label{sec:3.2}
\par As a more informative CD task, SCD can provide the detailed “from-to” change transition information for real applications \cite{Zheng2022}. However, in order to train an effective semantic change detector, high-quality land-cover labels of pre-event and post-event images and the binary change labels between them are required, greatly increasing the cost of this task \cite{Yang2022Asymmetric, Tian2022Large}. An appealing aspect of the task in this study is that we do not need to manually annotate land-cover labels of optical high-resolution images for SCD as well. This is because we can naturally go from the OSM data itself to the land-cover labels it represents \cite{SCHULTZ2017Open}. For the unchanged areas, the land-cover categories are consistent between OSM data and optical images. Therefore, we can perform a masking operation on the semantic labels derived from OSM data using binary change labels to obtain partial semantic change labels for the optical images. Then, we can train a semantic change detector based on them. This can be seen as a special semi-supervised learning task. Formally, given the training set $\mathcal{D}^{scd}_{train}=\left\{\left(\mathbf{X}^{osm}_{i}, \mathbf{X}^{opt}_{i}, \mathbf{Y}^{osm}_{i}, \hat{\mathbf{Y}}^{opt}_{i}, \mathbf{Y}^{bcd}_{i}\right)\right\}_{i=1}^{N_{train}}$, where $\mathbf{Y}^{osm}_{i}\in \left\{1,\cdots,\mathrm{C}^{lcm}\right\}^{\mathrm{H} \times \mathrm{W}}$ is the land-cover labels of OSM data and $\hat{\mathbf{Y}}^{opt}_{i}=\left(1-\mathbf{Y}^{bcd}_{i}\right)\odot \mathbf{Y}^{osm}_{i}\in \left\{0,1,\cdots,\mathrm{C}^{lcm} \right\}^{\mathrm{H} \times \mathrm{W}}$ is the partial land-cover labels of their optical imagery generated from $\mathbf{Y}^{osm}_{i}$ based on their consistency in unchanged areas, the goal is to train a detector $\mathcal{T}^{scd}$ that can generate land-cover maps of optical images and binary change maps between OSM data and optical images from new sets as accurately as possible. 

\begin{figure*}[!t]
  \centering
\includegraphics[width=6.8in]{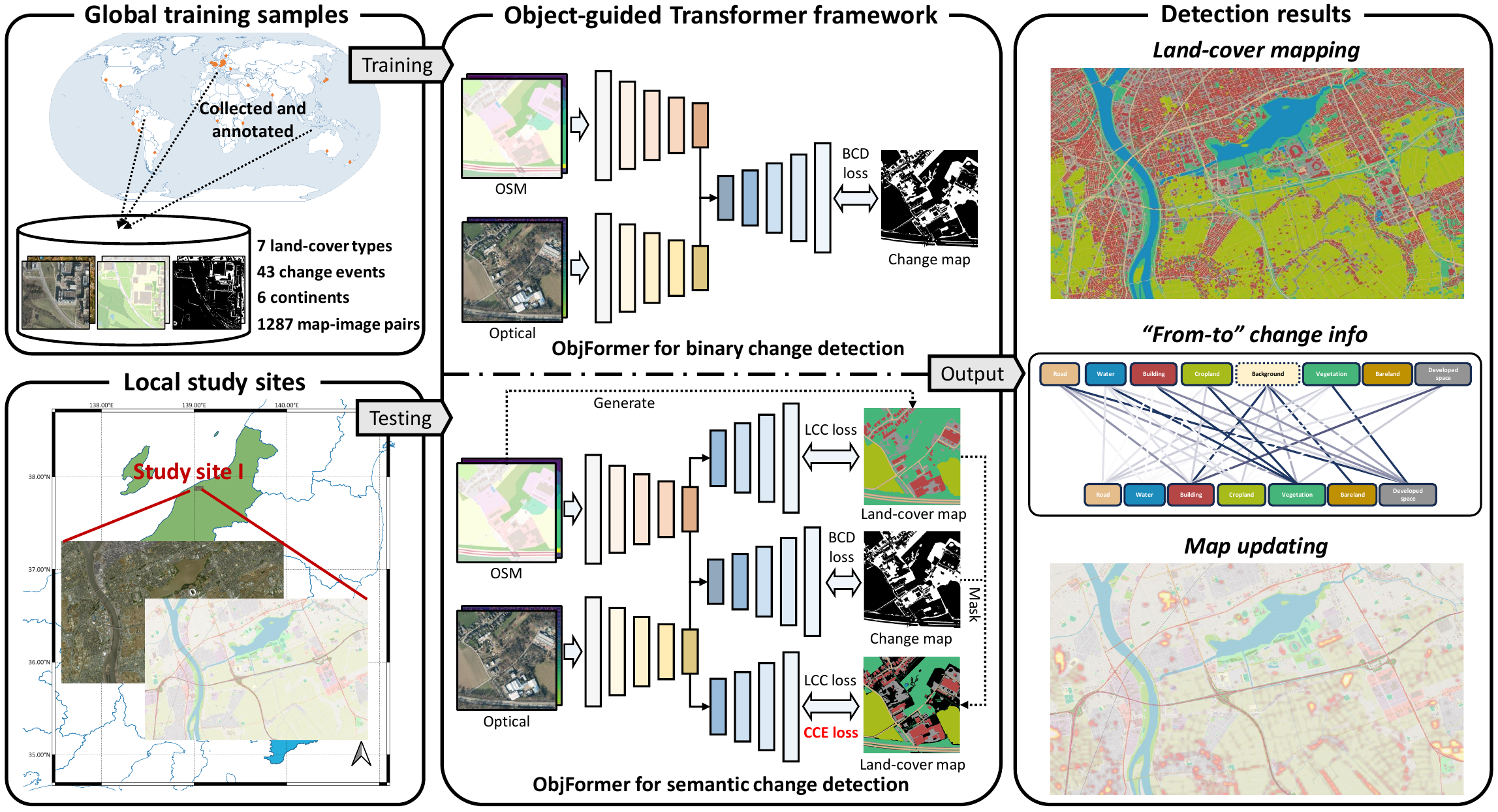}
  \caption{The overall workflow of CD on OSM data and optical high-resolution imagery based on the proposed ObjFormer architecture. }
  \label{fig:overall_workflow}
\end{figure*}

\subsection{Combining OBIA with Self-Attention}\label{sec:3.2}

\par In this subsection, we introduce how the OBIA technique is seamlessly incorporated into the Transformer architecture. Specifically, given an optical high-resolution image $\mathbf{X}^{opt}_{i}$, the segmentation algorithm is performed on it to obtain the object map $\mathbf{\Omega}^{opt}_{i}$. We utilize the formulation in \cite{Sun2021a, Chen2023Fourier} to express the object map as 

\begin{equation}
\left\{
    \begin{aligned}
      &\mathbf{\Omega}^{opt}_{i}=\{\mathbf{\Omega}^{opt}_{i}\left(n\right) \mid n=1,2,\ldots, N^{obj}_{i}\}  \\
      &\mathbf{\Omega}^{opt}_{i}\left(n\right)\cap\mathbf{\Omega}^{opt}_{i}\left(m\right)=\varnothing \ \ \mathrm{if} \ n\ \neq m  \\
      &\bigcup_{n=1}^{N^{obj}_{i}}\mathbf{\Omega}^{opt}_{i}\left(n\right)=\{\left(h,w\right)\mid h=1,\dots,\mathrm{H}; w=1,\ldots,\mathrm{W}\} 
    \end{aligned}
\right.
\label{eq:1}
\end{equation}
where $N^{obj}_{i}$ is the number of objects and $\mathbf{\Omega}^{opt}_{i}\left(n\right)$ is the $n$-th object. 

\par Then, denote the feature map of $i$-th sample in the $l$-th intermediate layer of a Transformer architecture as $\mathbf{F}_{i,l}\in \mathbb{R}^{\mathrm{H}_{l}\times \mathrm{W}_{l}\times \mathrm{C}_{l}}$, where $\mathrm{H}_{l}$, $\mathrm{W}_{l}$ and $\mathrm{C}_{l}$ are the height, width and channel number of $\mathbf{F}_{i,l}$, respectively. It is required to serialize it into individual tokens for the following self-attention module. The vanilla vision Transformer just reshapes $\mathbf{F}_{i,l}$ and treats each location as an individual token, i.e., $\mathbf{T}_{i,l}=\left[\mathbf{F}_{i,l}\left(1,1\right), \mathbf{F}_{i,l}\left(1,2\right), \cdots, \mathbf{F}_{i,l}\left(\mathrm{H}_{l}, \mathrm{W}_{l}\right)\right]^{T} \in \mathbb{R}^{\mathrm{H}_{l} \mathrm{W}_{l}\times \mathrm{C}_{l}}$. Here, instead of treating each location as an individual token, we downsample the object map $\mathbf{\Omega}^{opt}_{i}$ using the nearest neighbor algorithm to the same resolution as $\mathbf{F}_{i,l}$, and fuse all pixels inside each object to get the deep object tokens before the self-attention module, formulated as 

\begin{equation}
\left\{
    \begin{aligned}
      &\mathbf{T}^{obj}_{i,l}=\left[\mathbf{T}^{obj}_{l}(1), \mathbf{T}^{obj}_{l}(2), \cdots, \mathbf{T}^{obj}_{l}(N^{obj}_{i,l})\right]^{T} \in \mathbb{R}^{N^{obj}_{i,l}\times \mathrm{C}_{l}}  \\
      &\mathbf{T}^{obj}_{i,l}\left(n \right)=\mathcal{F}\left(\mathbf{F}_{i,l}\left(i,j\right)\right | \left(i,j\right)\in \mathbf{\Omega}^{opt}_{i,l}\left(n\right) )
    \end{aligned}
\right.
\label{eq:2}
\end{equation}
where $\mathbf{T}^{obj}_{i,l}$ is deep object tokens, $\mathbf{\Omega}^{opt}_{i,l}$ represents the downsampled object map, $N^{obj}_{i,l}$ is the number of objects $\mathbf{\Omega}^{opt}_{i,l}$ contains, and $\mathcal{F}(\cdot)$ represents a feature fusion method. For the purpose of not altering the network structure or introducing new parameters, we use non-parametric fusion methods, such as taking the mean value of pixels within the object on each channel. The effect of different fusion methods will be discussed in Section \ref{sec:effect_OBIA}.

\par By the above means, the computational complexity of the self-attention mechanism is significantly reduced from $O\left(\mathrm{H}_{l}\mathrm{W}_{l} \times \mathrm{H}_{l}\mathrm{W}_{l}\right)$ to $O\left(N^{obj}_{l} \times N^{obj}_{l}\right)$, where $N^{obj}_{l}<<\mathrm{H}_{l}\mathrm{W}_{l}$, and the size of the self-attention map used for forward and backward propagation is also reduced by a corresponding order of magnitude. Such an object-guided self-attention method combines the OBIA technique and self-attention mechanism in a simple and natural way without adding new modules or modifying the network architecture to reduce the computational and memory burden of the self-attention mechanism. Since each object represents a homogeneous region, reducing the number of tokens in this way can considerably preserve the information of the feature map. Moreover, the above process does not introduce new trainable parameters and can be seamlessly integrated into most off-the-shelf vision Transformer architectures \cite{dosovitskiy2020image,Wang2021Pyramid, Xie2021SegFormer}. 

\par For the OSM data, we make targeted improvements to the object-guided self-attention method. We can easily get individual instances at higher semantic levels in the OSM data, not just homogeneous localized regions in the case of optical imagery. Thus, for an OSM data sample $\mathbf{X}^{osm}_{i}$, instead of generating a corresponding object map, we go a step further to generate an instance map $\mathbf{\Omega}^{osm}_{i}=\{\mathbf{\Omega}^{osm}_{i}(n) \mid n=1,2,\ldots, N^{ins}_{i}\}$, where $N^{ins}_{i}$ is the number of instances in the OSM data, generally $N^{ins}_{i}<< N^{obj}_{i}$. As a result, when processing the OSM data, the number of generated tokens is quite few, thereby reducing the computational and memory burden of the self-attention mechanism even further compared to the case of processing optical imagery.

\begin{figure*}[!t]
  \centering
\includegraphics[width=6.3in]{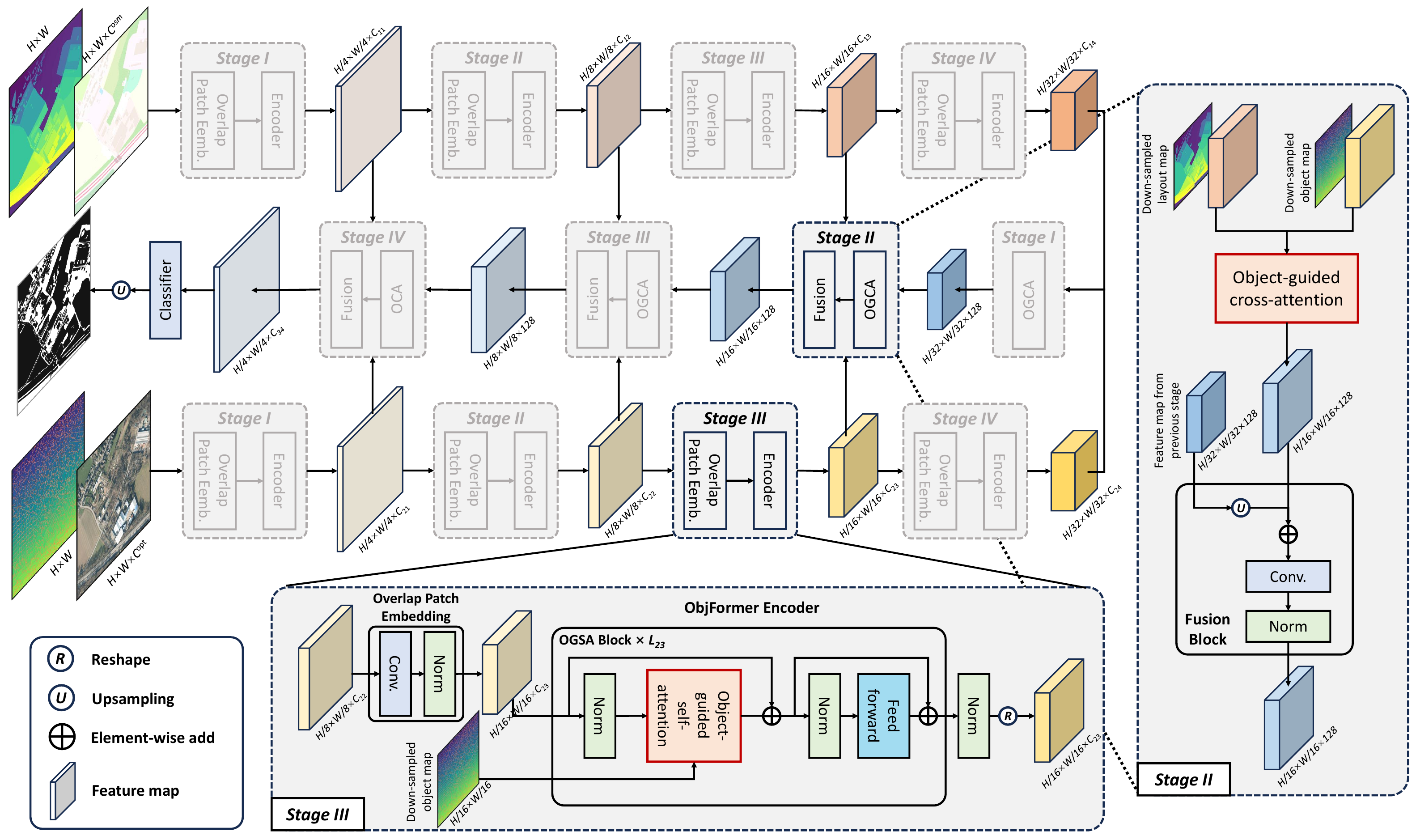}
  \caption{Network architecture of the proposed ObjFormer for BCD on OSM data and optical remote sensing images. }
  \label{fig:ObjFormer_arc}
\end{figure*}

\par As for the choice of a specific segmentation algorithm, there are many approaches available \cite{Achanta2012SLIC, Comaniciu2002Mean}. It is beyond this paper's scope to investigate which is the best for CD on paired OSM data and optical images. In this study, we used the simple linear iterative clustering (SLIC) algorithm \cite{Achanta2012SLIC} implemented by the scikit-learn library\footnote{https://scikit-learn.org/} to generate object maps of optical images. This is because, in the implementation of this algorithm in scikit-learn, there is a hyperparameter called $\left[n\_segments\right]$, allowing us to easily control the number (or say scale) of the generated object for the subsequent analysis and discussion. 

\subsection{Object-Guided Transformer}\label{sec:3.3}
\par Leveraging the proposed object-guided self-attention mechanism, we present a vision Transformer architecture called ObjFormer for CD on paired OSM data and optical imagery. The overall workflow is shown in Fig. \ref{fig:overall_workflow}. Two network architectures are proposed for BCD and SCD tasks, respectively. Fig. \ref{fig:ObjFormer_arc} illustrates the architecture for BCD, consisting of a pseudo-siamese hierarchical encoder for extracting multi-level features and a heterogeneous information fusion decoder to interpret land-cover changes. Two simple semantic decoders are added to the BCD architecture to accomplish the SCD task.

\subsubsection{\textbf{Pseudo-siamese hierarchical encoder}}
\par The vanilla vision Transformer is proposed for the image classification task \cite{dosovitskiy2020image}, where the feature map resolution keeps a single low resolution during the feature extraction process. However, for dense visual prediction tasks, including CD, the design of the encoder for extracting features from high-resolution and low-level to low-resolution and high-level is necessary for accurate prediction results \cite{Wang2021Pyramid}. Due to the large domain gaps between the OSM data and the optical images, the encoder of our network is designed as a pseudo-siamese structure, where the two branches process their input separately and do not share parameters. As shown in Fig. \ref{fig:ObjFormer_arc}, each branch has four stages, and each stage includes an overlapping patch embedding module and an object-guided Transformer-based encoder. 

\begin{figure}[!t]
  \centering
\includegraphics[width=3.4in]{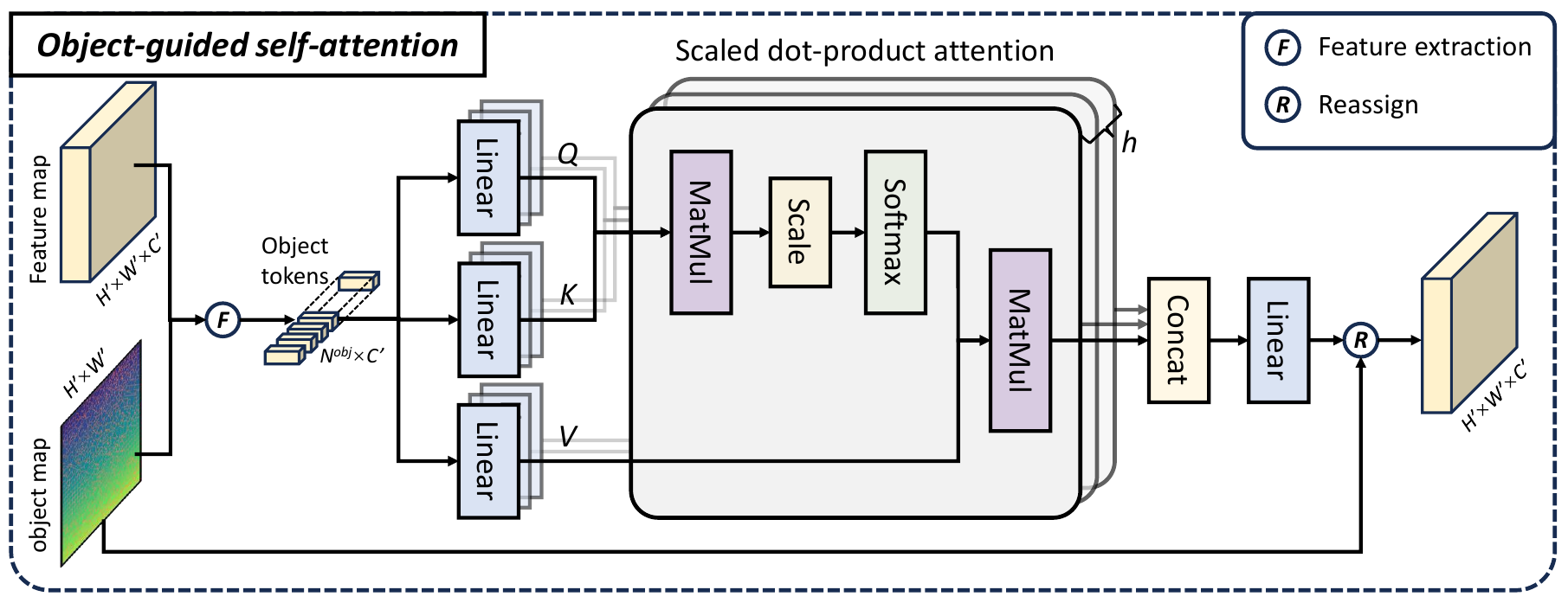}
  \caption{Detailed structure of the proposed object-guided self-attention module. Here, reassign means to reconvert the tokens modeled by the self-attention mechanism into a feature map based on the coordinate information provided by the object/instance map.}
  \label{fig:OSA_module}
\end{figure}

\par The overlap patch embedding module first shrinks the feature map to generate embeddings, allowing the whole encoder to generate multi-level feature maps for downstream tasks. Specifically, this module utilizes a stride convolutional layer to shrink the size of feature maps. This can preserve the local continuity around those patches compared to non-overlap patch embedding \cite{Xie2021SegFormer}. Following the overlap patch embedding, the object-guided Transformer encoder extracts representative features from the input. After normalized by a layer normalization module \cite{ba2016layer}, the feature is encoded by our object-guided self-attention module, with the structure depicted in Fig. \ref{fig:OSA_module}. Similar to the original self-attention \cite{Vaswani2017Attention}, this module receives a query $Q$, a key $K$, and a value $V$ as input and outputs a refined feature map. The difference is that object/instance tokens are first generated based on the object/instance map, as described in Section \ref{sec:3.2}, to greatly reduce the sequence length and then undergo transformation through linear layers to yield $Q$, $K$, and $V$. Then, the standard self-attention mechanism is performed to get refined features. After that, the features will be processed by a layer normalization module and a feedforward layer. Here, the feedforward layer contains a 3$\times$3 convolutional layer to implicitly encode the positional information into the generated features \cite{islam2020much, Xie2021SegFormer}, since the self-attention mechanism does not take the local information into account. Compared to the vanilla positional encoding in \cite{dosovitskiy2020image}, which can only handle fixed-size inputs, this way can allow our network to accommodate input of variable size. The above process will be repeated $L_{i}$ times in stage $i$, and finally performed on the output features. 
Through the four stages, each branch yields four feature maps with different resolutions and levels of semantic information. 

\par It is worth noting that the branch processing OSM data is designed to be more lightweight than that of processing optical images, focusing primarily on learning mappings from specific colors to land-cover categories and understanding the spatial distribution of different instances. This design choice, which includes a reduced number of channels and modules, leverages the uniform value assigned to each category in the OSM data, optimizing the learning process.

\subsubsection{\textbf{Heterogeneous information fusion decoder}}

\par After extracting features from the input OSM data and optical images, the subsequent task is to interpret land-cover changes from these multi-level features. However, the features from the two branches are heterogeneous since these two data types have different modalities. Here we adopt the cross-attention mechanism \cite{Vaswani2017Attention}, which is a variant of self-attention and has shown promising results in fusing multimodal information like text and image data \cite{rombach2022high}, to fuse these two heterogeneous features. Naturally, we design an object-guided cross-attention module to greatly reduce the computational overhead in the original cross-attention module. 

\par
\begin{figure}[!t]
  \centering
\includegraphics[width=3.5in]{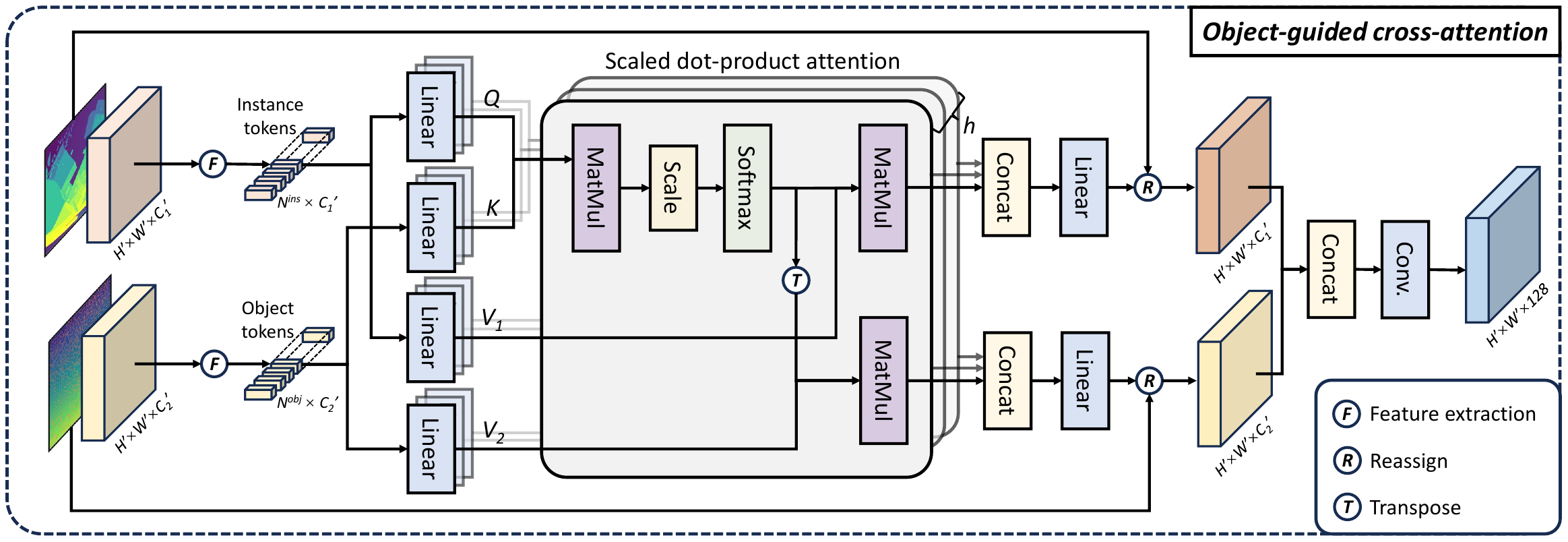}
  \caption{Detailed structure of the proposed object-guided cross-attention module for fusing heterogeneous features. }
  \label{fig:OGCA_module}
\end{figure}

 As shown in Fig. \ref{fig:OGCA_module}, object and instance tokens are first generated according to the associated object and instance maps. Then, cross-attention is performed on $Q$, $K$, $V^{osm}$, and $V^{opt}$, where $Q$ and $V^{osm}$ are derived from the tokens of the OSM data and $K$ and $V^{opt}$ are derived from that of optical imagery. Object-guided cross-attention is similar to self-attention. The only difference is that the $Q$ and $K$ for computing the multi-head self-attention maps are derived from object tokens and instance tokens, respectively, instead of one type of tokens. Then, we will multiply $V^{osm}$ and $V^{opt}$ with the calculated attention maps to get the two refined features. These two features will be used for subsequent SCD. For BCD, the two features are concatenated in the channel dimension and then fused through a point-wise convolutional layer.

\par The decoder also has four stages. In each stage except for the first one, the heterogeneous features of the same size from both branches are first fused through an object-guided cross-attention module, and then a fusion block upsamples the coarse-resolution fused features from the previous stage and merges it into the finer-resolution fused features with an element-wise addition operation and a 3$\times$3 convolutional layer for feature smoothing. Finally, the feature map with the finest resolution is generated after processed by the four stages. A 1$\times$1 convolutional layer is applied as the classifier to predict the binary change map.

\subsubsection{\textbf{Auxiliary semantic decoder}}

\par For SCD, we just incorporate two auxiliary semantic decoders into the architecture proposed for BCD \cite{Rodrigo2019Multitask, Zheng2022}. The two semantic decoders will predict land-cover classification maps of OSM data and optical imagery from the multi-level features extracted by their respective corresponding encoders. The semantic decoder in this work is composed of four fusion blocks. They have the same structure as the fusion blocks in the heterogeneous information fusion decoder. It upsamples and fuses the deep features generated by the object-guided cross-attention module to predict land-cover classification maps of OSM data or optical images. Compared to the basic architecture proposed for BCD, the auxiliary decoder is very lightweight and does not add many network parameters and computational overhead.

\subsection{Optimization}
\subsubsection{\textbf{BCD}}
\par As BCD can be seen as a kind of special semantic segmentation task, we directly apply cross-entropy loss to train the BCD architecture as

\begin{equation}
    \mathcal{L}_{bcd} = -\frac{1}{N_{train}} \sum_{i=1}^{N_{train}} \mathbf{Y}_{i}^{bcd}log(\mathbf{P}_{i}^{bcd}),
\end{equation}
where $\mathbf{P}^{bcd}_{i} \in \mathbb{R}^{\mathrm{H}\times \mathrm{W} \times 2}$ is the prediction map pertaining to $i$-th training sample.
\subsubsection{\textbf{SCD}}
\par  As for SCD, the network has two auxiliary decoders to predict the land-cover mapping results. They will also be trained using cross-entropy loss, denoted as $\mathcal{L}^{osm}_{lcm}$ and $\mathcal{L}^{opt}_{lcm}$. Despite the direct availability of semantic labels for OSM data, the decoder is still trained to predict the land-cover maps of OSM data to facilitate the network's overall learning process. As described in Section \ref{sec:3.1}, the SCD task we are concerned with in this paper is a semi-supervised learning task. The land-cover labels for the optical data are incomplete and only land-cover labels for the unchanged areas are available, which are generated from the OSM data.

\par During the process described above, the information in the changed areas is not utilized at all. Here, we propose a new loss function for such semi-supervised SCD, called converse cross-entropy (CCE) loss, which can effectively harness the information from the changed areas. The basic idea of this loss function is that although we do not know what the exact category of changed areas is, we can know what the exact category of changed areas is not. For example, for a given changed area, where the area is a water body in the OSM data, it must not be a water body in the corresponding area of the optical image. Such complementary information can contribute to the recognition tasks \cite{ishida2017learning}. According to this insight, the specific form of CCE loss is expressed as 

\begin{equation}
    \mathcal{L}_{cce} = -\frac{1}{N_{train}} \sum_{i=1}^{N_{train}} \mathbf{Y}_{i}^{bcd}\mathbf{Y}_{i}^{osm}log(1-\mathbf{P}_{i}^{opt}),
    \label{eq:cce_loss}
\end{equation}
where the probability of an unlikely category, i.e., the same category as that of the OSM data, will be minimized during the training phase, thus allowing the network to learn partial knowledge of the changed region. 

\par In summary, the final loss function for the SCD task is

\begin{equation}
\mathcal{L}_{scd}=\mathcal{L}_{bcd}+\mathcal{L}^{osm}_{lcm}+\mathcal{L}^{opt}_{lcm}+\mathcal{L}_{cce}. 
\end{equation}

\subsubsection{\textbf{Multi-scale object learning}}
\par Sections \ref{sec:3.2} and \ref{sec:3.3} show that OBIA and the specific network architecture are actually decoupled in our proposed object-guided Transformer. Therefore, the scale of the object can be adjusted flexibly. Based on this attribute, a multi-scale object learning approach is proposed. We first generate several object maps with different scales for each optical imagery. Then, these object maps of varying scales are randomly selected as the input in each iteration of the training phase. Subsequently, the results derived from multiple object maps are fused in the testing phase, enhancing the robustness and accuracy of the predictions. This approach also avoids the determination of the optimal scale of object maps in the single-scale case. 

\section{Experimental setup}\label{sec:4}
\subsection{Implementation Details}
\par The proposed CD architecture was implemented using PyTorch. For the specific network architecture, the block number of the four stages in the map branch is $\left[2, 2, 2, 2\right]$ with the channel number of the obtained feature maps of $\left[32, 64, 160, 256\right]$; the block number of the four stages in the optical branch is $\left[3, 4, 6, 3\right]$ with the channel number of the feature maps of $\left[64, 128, 320, 512\right]$. The downsampling factor of the four overlap patch embedding modules is $\left[1/4, 1/2, 1/2, 1/2\right]$. The number of heads of the self-attention modules in the four stages of the encoder is [1, 2, 5, 8]. Corresponding to the number of heads in the encoder, the number of heads of the cross-attention module in the decoder is [8, 5, 2, 1]. AdamW optimizer \cite{loshchilov2017decoupled} was utilized for training the networks with a batch size of 16, a learning rate of 1$e^{-4}$, weight decay of 5$e^{-3}$, and total iterations of 7500 and 10000 for binary and SCD tasks, respectively. The random crop with a patch size of 512$\times$512, random horizontal and vertical flips, and random rotation were applied for training data augmentation. For OBIA, the scikit-learn was used to implement the SLIC algorithm to obtain object maps of optical high-resolution images and the connected component labeling algorithm to obtain instance maps of OSM data. All experiments were conducted with an NVIDIA A100 Tensor Core GPU. The source code of ObjFormer will be open-sourced for replication and subsequent research of the community\footnote{The source code of this work will be available at: https://github.com/ChenHongruixuan/ObjFormer}. 

\subsection{Evaluation Metrics}\label{sec:eval_metrics}
\par For BCD on paired OSM data and optical images, we adopt five commonly used accuracy metrics to evaluate the performance of different detectors \cite{CAO2023full}. They are the recall rate (Rec), precision rate (Pre), F1 score, overall accuracy (OA), and Kappa coefficient (KC). 

\par For SCD, we need to simultaneously show the performance of the detectors in both land-cover mapping and CD tasks \cite{Wu2017b}. For the accuracy of land-cover mapping results on the optical images, we adopt OA and KC as the evaluation metrics. To differentiate them from the indices in the above BCD task, we denote them as clfOA and clfKC, respectively. For CD, we use KC to evaluate the accuracy of binary change maps obtained by detectors as well as OA and KC to evaluate the accuracy of the final semantic change maps providing “from-to” change transition information, which are denoted as cdKC, trOA, and trKC, respectively. 

\par In addition, we adopt the number of trainable parameters and multiply-accumulate operations (MACs) to quantitively measure the models' complexity and computational overhead. 

\begin{figure*}[!ht]
  \centering
\includegraphics[width=7.0in]{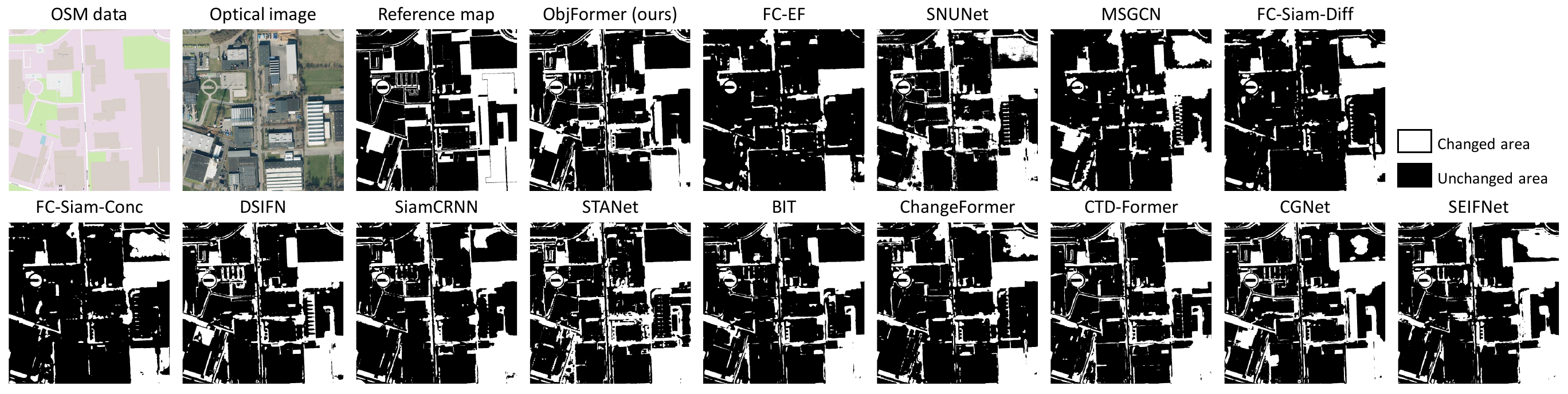}
  \caption{Binary land-cover change maps obtained by different methods on a test sample in Aachen, German.}
  \label{bcm_aachen_4}
\end{figure*}

\subsection{Benchmark methods}\label{sec:benchmark_methods}
\par Since our work pioneers detecting land-cover changes directly from paired OSM data and optical imagery, our comparison methods can only be selected from those already proposed for homogeneous and heterogeneous CD. We chose 13 representative models for the BCD task as comparison methods. These approaches encompass the various deep architectures currently dominant for CD, i.e., CNN, RNN, GCN, Transformer and their combinations, as well as two design paradigms for change detector structures, i.e., early fusion structures and siamese structures.
They are FC-EF \cite{CayeDaudt2018}, SNUNet \cite{Fang2022SNUNet}, MSGCN \cite{Wu2021multiscale}, FC-Siam-Diff \cite{CayeDaudt2018}, FC-Siam-Conc \cite{CayeDaudt2018}, DSIFN \cite{Zhang2020}, SiamCRNN \cite{Chen2019a}, STANet \cite{Chen2020Spatial}, BIT \cite{Chen2022Remote}, and ChangeFormer \cite{Bandara2022Transformer}, CTD-Former \cite{Zhang2023Relation}, CGNet \cite{Han2023Change}, and SEIFNet \cite{Huang2024Spatiotemporal}.

\par Compared to BCD, there is relatively little work on SCD. Here, eight representative approaches are adopted as comparison methods. They are HRSCD-S1 \cite{Rodrigo2019Multitask}, HRSCD-S3 \cite{Rodrigo2019Multitask}, HRSCD-S4 \cite{Rodrigo2019Multitask}, ChangeMask \cite{Zheng2022}, SSCD-1 \cite{Ding2022Temporal}, BiSRNet \cite{Ding2022Temporal}, TED \cite{Ding2024Joint}, and SCanNet \cite{Ding2024Joint}. 
\par Note that to make the above methods employing a siamese network structure, initially designed for homogeneous images, suitable for processing heterogeneous OSM data and optical images, we modify their network structure to a pseudo-siamese structure. 

\begin{figure*}[!ht]
  \centering
\includegraphics[width=7.0in]{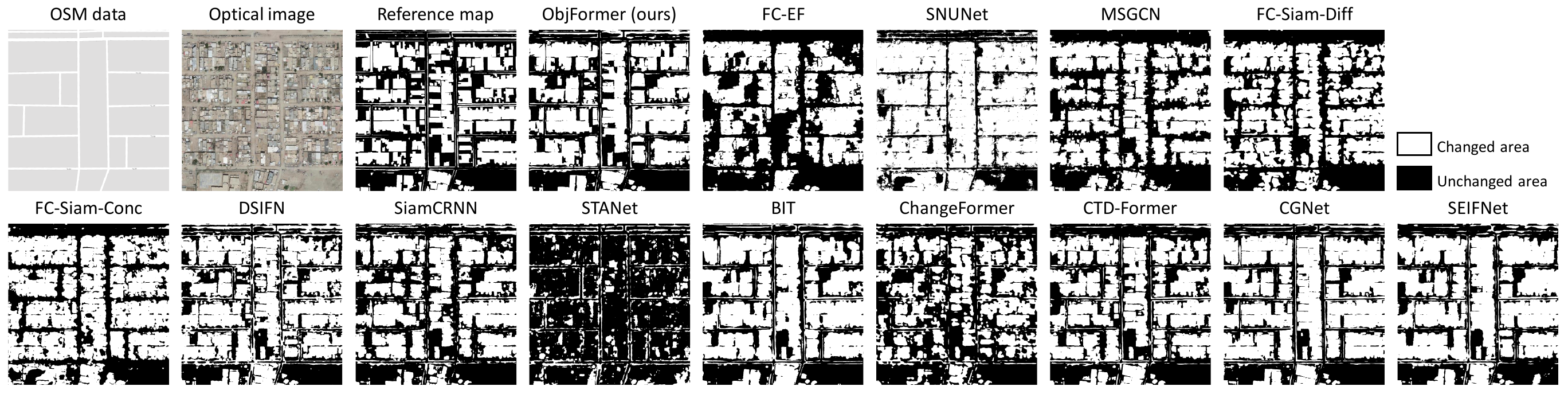}
  \caption{Binary land-cover change maps obtained by different methods on a map-image test sample in Al-Qurnah, Iraq.}
  \label{bcm_al_qurnah_19}
\end{figure*}

\begin{table}[!t]
  \renewcommand{\arraystretch}{1.3}
\caption{\centering{Accuracy assessment for different BCD models on the OpenMapCD dataset. The table highlights the highest values in bold, and the second-highest results are underlined.}}\label{tbl:acc_bcd}
  \centering
  \begin{tabular}{c c c c c c}
  \toprule
    \hline	
    \textbf{Method}	&	\textbf{Rec}	& \textbf{Pre} 	 &  \textbf{OA}  & \textbf{F1}    & \textbf{KC} \\
    \hline\hline
   FC-EF &	0.7008 &	0.7699 &	0.8128 &	0.7337 &	0.5898 \\
SNUNet &	0.8296	 &0.6891	 &0.7995	 &0.7528	 &	0.5867 \\
MSGCN &	0.7783	 &0.8186	 &0.8549 &	0.7979	 &	0.6849 \\

FC-Siam-Diff &	0.6983 &	0.8029 &	0.8259 &	0.7469	 &	0.6152 \\
FC-Siam-Conc &	0.7442	 &0.7608 &	0.8197	 &0.7524	 &0.6107 \\
DSIFN	 &0.8344	 &0.8544 &	0.8867	 &0.8443	 &	0.7553 \\
SiamCRNN &	0.8374	 & 0.8622 &	0.8909	 & 0.8496	 &	0.7640 \\
STANet	 &0.7252	 &0.7474 &	0.8086	 &0.7361  &	0.5861 \\
BIT &	0.8359	 &0.8463 &	0.8837 &	0.8411	 &	0.7495 \\
ChangeFormer &	0.8060	 &0.7857	 & 0.8477 &	0.7957	 &	0.6743 \\
CTD-Former &	0.8090 & 0.8239	 &  0.8661 & 0.8164	 & 0.7110 \\
CGNet &	\underline{0.8697} & 0.8469 & \underline{0.8942} & \underline{0.8581}	 & \underline{0.7738} \\
SEIFNet &	0.8305 & \underline{0.8713}	 & 0.8925 & 	0.8504 & 0.7666 \\
ObjFormer &	\textbf{0.8729}	 & \textbf{0.8812}	 & \textbf{0.9099}	 & \textbf{0.8770}	 &	\textbf{0.8059} \\

    \hline
    \bottomrule
  \end{tabular}
\end{table}

\begin{table}[!t]
  \renewcommand{\arraystretch}{1.3}
\caption{\centering{Network parameters and computational overhead of different BCD models with the input size of 512$\times$512.}}\label{tbl:param_mac} 
  \centering
  \begin{tabular}{c c c}
  \toprule
    \hline	
   \textbf{Method}	& \textbf{Parameters (M)}	& \textbf{MACs (G)}  \\
    \hline\hline
   FC-EF &	1.35	 &	14.13 \\
    SNUNet&	11.38 &	176.36 \\
    MSGCN&	22.23 &	117.58 \\
    FC-Siam-Diff&	1.83 &	18.66 \\
    FC-Siam-Conc&	2.03 &	21.07 \\
    DSIFN&	51.23 &	329.03 \\
    SiamCRNN&	49.80 &	110.09 \\
    STANet&	24.38 &	114.53  \\
    BIT&	5.99 &	34.93 \\
    ChangeFormer &	48.50	& 52.33 \\
    CTD-Former &  6.71 &  303.77 \\
    CGNet & 48.40 & 329.28  \\
    SEIFNet & 39.08 & 167.75 \\
    ObjFormer &	28.37 &	27.12\\
    \hline
    \bottomrule
  \end{tabular}
\end{table}

\section{Experimental results and discussions}\label{sec:5}
\subsection{Benchmark Comparison and Analysis}
\par We first investigate and analyze the performance of our framework and the existing representative methods proposed for homogeneous and heterogeneous CD tasks. To fairly evaluate these methods, we train all these comparison methods using the same protocol as our approach and then evaluate their detection performance using the accuracy metrics in Section \ref{sec:eval_metrics} on the test set. On the BCD task, we choose the iteration with the highest KC as the best model. On the SCD task, we choose the iteration with the highest trKC as the best model.

\subsubsection{BCD}

\par Fig. \ref{bcm_aachen_4} and Fig. \ref{bcm_al_qurnah_19} show the binary change maps obtained by our method and the 13 comparison methods on the two map-image pairs. In the first example, these methods yield visually appealing change maps. However, the data forms of OSM data and optical images are different and the domain gap between them is large. As a result, many comparison methods still have misdetections and omissions in some local areas. For example, SNUNet, MSGCN, FC-Siam-Diff, FC-Siam-Conc, ChangeFormer, CGNet, and SEIFNet incorrectly detect the vegetation area in the upper right region as a changed area. In addition, almost all methods fail to detect the newly constructed building in the lower left. In contrast, our method can extract representative features using two Transformer-based branches and then utilize the object-guided cross-attention modules to effectively fuse heterogeneous features, resulting in a change map visually closest to the reference map. 

\par In the second example, the buildings are the main difference between the OSM data and the optical high-resolution image. None of the built-up areas in the OSM data are edited or drawn. Thus, this example could exemplify the potential of different approaches for directly updating buildings for map data. From Fig. \ref{bcm_al_qurnah_19}, we can see that the detection results of these comparison methods either show over-smoothing or are not complete enough with internal fragmentation. Also, the metric learning-based approach, STANet, does not work well in this case, barely detecting changed areas. In contrast, combining the advantages of the OBIA method and the Transformer architecture, our ObjFormer detects visually the best change map that is closest to the reference map, with clear boundaries and complete change regions.

\par The overall quantitative results of our framework and comparison methods are reported in Table \ref{tbl:acc_bcd}. We also list their number of trainable parameters and MACs in Table \ref{tbl:param_mac}. The first conclusion we can draw is that the pseudo-siamese structure can achieve better detection results on the task of CD using paird OSM data and optical imagery compared to the early-fusion structure. For example, the benchmark methods FC-Siam-Diff and FC-Siam-Conc outperform FC-EF. The early-fusion SOTA method SNUNet, achieving decent performance on several homogeneous datasets \cite{Fang2022SNUNet}, can only achieve KC values similar to FC-EF on our task. Some siamese SOTA methods, e.g., CGNet and SEIFNet, perform significantly better than SNUNet and MSGCN. This is because the pseudo-siamese structure can directly extract representative features from input data belonging to different domains, which cannot be done explicitly with the early-fusion structure. The second conclusion is that the fusion of heterogeneous features is important in the task of this paper. The five advanced methods, DSIFN, SiamCRNN, BIT, CGNet, and SEIFNet, which achieve higher accuracy than other methods, use the attention mechanism, multiple-layer ConvLSTM, and Transformer to fuse extracted heterogeneous features, respectively. In contrast, ChangeFormer, another SOTA method based on the Transformer backbone, only achieves a KC value of 0.6743. This is because its decoder directly performs element-wise subtraction on the extracted heterogeneous features, which is only suitable for homogeneous datasets with small domain gaps. 

\par Finally, our method achieves the highest accuracy on all five metrics, validating the superiority of our method on this task. From Table \ref{tbl:param_mac}, it can be found that compared to SOTA methods such as DSIFN, CGNet, and SEIFNet, our method has fewer network parameters and lower MACs benefiting from the design of asymmetric encoder branches and the introduction of the OBIA method to reduce computational overhead.

\subsubsection{\textbf{SCD}}

\begin{figure*}[!t]
  \centering
\includegraphics[width=7.0in]{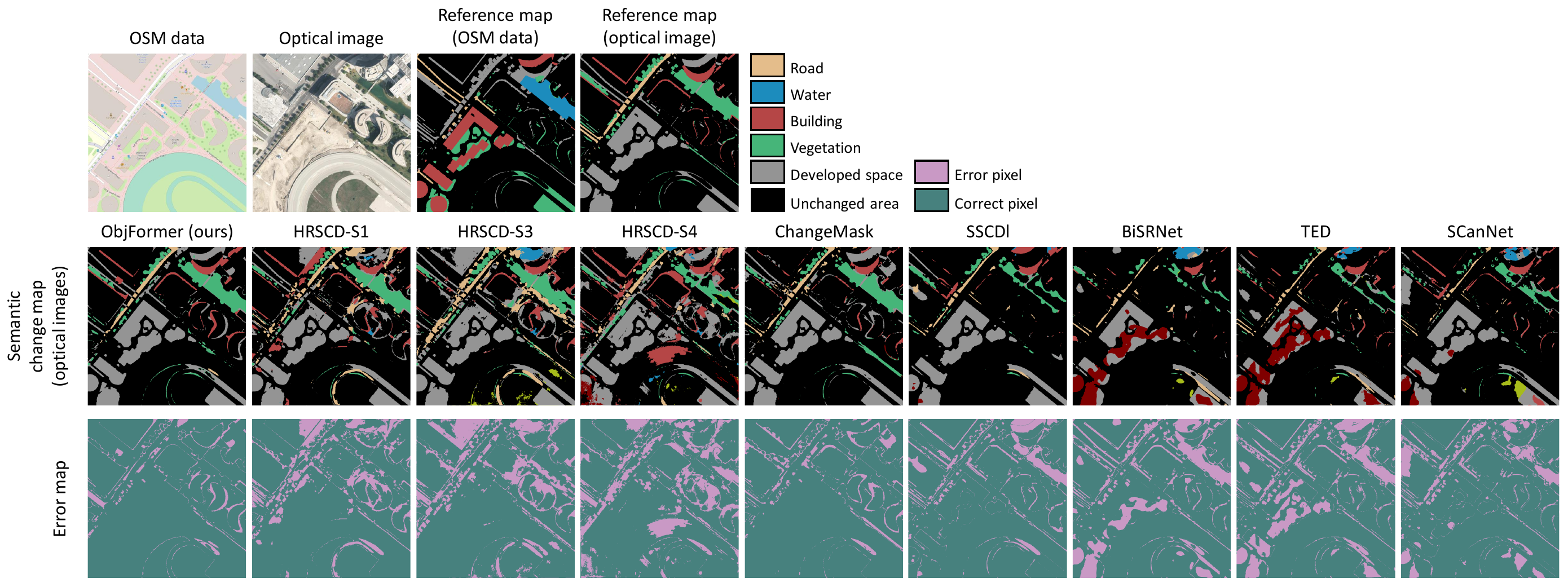}
  \caption{Semantic land-cover change maps and associated error maps of different methods on a test sample in Vienna, Austria.}
  \label{scm_vienna_8}
\end{figure*}

\begin{figure*}[!t]
  \centering
\includegraphics[width=7.0in]{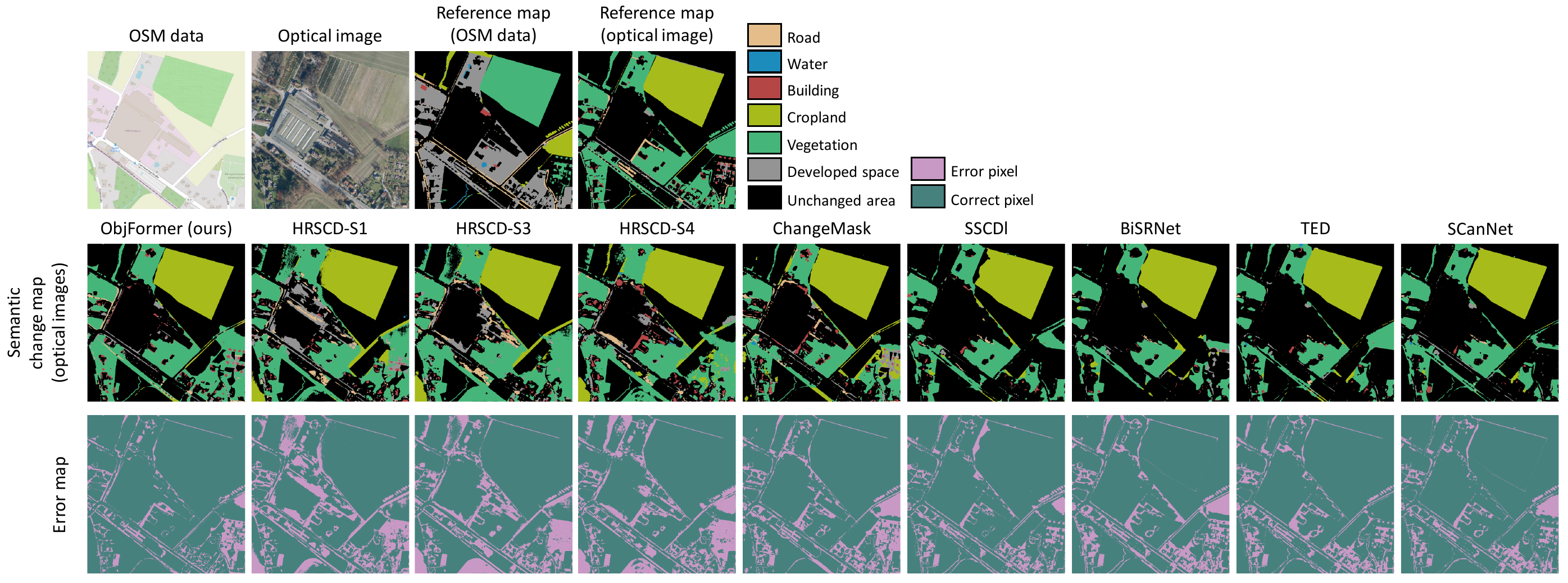}
  \caption{Semantic land-cover change maps and associated error maps of different methods on a test sample in Bielefeld, German.}
  \label{scm_Bielefeld_10}
\end{figure*}

\par Due to category combinations, SCD often involves a considerable number of categories. In our case, the number is 50. For ease of visualization, we follow the visualization way in previous studies \cite{Zheng2022, Tian2022Large, Yang2022Asymmetric}. That is, only plotting the land-cover categories of the changed regions in both pre- and post-change images (in our case, they are OSM data and optical images), which can reflect both binary “changed-unchanged” information and semantic “from-to” transition information. Moreover, we additionally draw error maps to facilitate visual comparisons. 

\par Fig. \ref{scm_vienna_8} and Fig. \ref{scm_Bielefeld_10} show two examples of SCD obtained by ObjFormer and comparison methods on the test set. The first example covers an urban scene. The major semantic changes are “developed space to buildings”. The semantic change maps obtained by several comparison methods have many misclassification pixels. By comparison, benefiting from the targeted network architecture and the utilization of CCE loss, our method yields the least errors between the semantic change map and the reference map, with changed regions clearly recognizable from each other and a high degree of completeness within changed regions. The second example shows a rural region. The major semantic changes are “vegetation to cropland” and “developed space to vegetation”. In this example, ObjFormer also obtains the most accurate semantic change map, although a small portion of the “farmland to vegetation” changes are not detected in the right area. 

\begin{table}[!t]
  \renewcommand{\arraystretch}{1.3}
\caption{\centering{Accuracy assessment for different SCD models on the OpenMapCD dataset. The table highlights the highest values in \textbf{bold}, and the second-highest results are \underline{underlined}. Further experiments on the proposed CCE loss are presented in Section \ref{sec:effect_cce}.}}\label{tbl:acc_scd} 
  \centering
  \begin{tabular}{c c c c c c}
  \toprule
    \hline	
    \textbf{Method}	& \textbf{clfOA}	& \textbf{clfKC} &	\textbf{cdKC} &	\textbf{trOA} &	\textbf{trKC}  \\
    \hline\hline
HRSCD-S1&	0.6869&	0.6006	& 0.4177	& 0.7089&	0.4558 \\
HRSCD-S3 &	0.6877&	0.5999&	0.6127 &	0.7551&	0.4847 \\
HRSCD-S4& 	0.6839&	0.5964& 0.6734 &	0.7570 &	0.5044 \\
ChangeMask &	\underline{0.7538} &	\underline{0.6852} &	0.6690  &\underline{0.7793}	&	0.5356 \\
SSCD-1 &	0.7372  &	0.6639  & 0.7159  & 0.7710 &	0.5205  \\
BiSRNet & 0.7322 & 0.6593 & 0.6951	& 0.7743 & 0.5364 \\
TED &	0.7426  &	0.6733  & 0.6620 & 0.7722 &	0.5237  \\
SCanNet &	0.7460  &	0.6750  & \underline{0.7356} & 0.7781 & \underline{0.5427}  \\
ObjFormer	& \textbf{0.8375} &	\textbf{0.7899} & \textbf{0.7896} &	\textbf{0.8740} &	\textbf{0.7654} \\
    \hline
    \bottomrule
  \end{tabular}
\end{table}

\begin{table}[!t]
  \renewcommand{\arraystretch}{1.3}
\caption{\centering{Network parameters and computational overhead of different SCD models with the input size of 512$\times$512.}}\label{tbl:param_mac_scd} 
  \centering
  \begin{tabular}{c c c}
  \toprule
    \hline	
   \textbf{Method}	& \textbf{Parameters (M)}	& \textbf{MACs (G)}  \\
    \hline\hline
HRSCD-S1 &	4.93 &	9.21 \\
HRSCD-S3&	9.86 &	18.44 \\
HRSCD-S4&	13.29 &	27.64\\
ChangeMask&	11.33 &	15.48 \\
SSCD-1 & 44.66 & 189.76 \\
BiSRNet &	 44.73 & 190.30 \\
TED & 45.54 & 204.29 \\
SCanNet &	49.25  & 264.95	  \\
ObjFormer & 31.57 &	33.16\\
    \hline
    \bottomrule
  \end{tabular}
\end{table}

\par Table \ref{tbl:acc_scd} reports the overall quantitative results of these methods on the test set. Table \ref{tbl:param_mac_scd} lists their number of trainable parameters and MACs. First, compared to the remaining eight methods, HRSCD-S1, which is based on the PCC, achieves the lowest trKC, which is in line with the conclusions of previous studies \cite{Rodrigo2019Multitask, Zheng2022}, even though the land-cover maps of the OSM data in our task are considered to be entirely accurate. By decoupling SCD into a land-cover classification task and a BCD task, HRSCD-S3 can achieve better detection results, with a 2.89$\%$ improvement in trKC compared to HRSCD-S1. Then, HRSCD-S4 further improves cdKC by 6.07$\%$ and trKC by 1.97$\%$ on the basis of HRSCD-S3, demonstrating that the multi-task learning can facilitate BCD in the context of SCD on paired OSM data and optical imagery. SCanNet achieves the second-highest accuracy on the SCD task by using Transfomer for joint spatio-temporal modeling. On the five evaluation metrics, our method achieves the best values. Especially in the final SCD result, ObjFormer outperforms the SOTA architecture SCanNet by 22.27$\%$ in trKC. 

\par These results indicate that ObjFormer can not only detect the changed-unchanged regions well but also accurately distinguish the specific “from-to” transition information. Such superior detection results come in part from the fact that CCE loss makes full utilization of information in negative samples. We will further demonstrate this point in Section \ref{sec:effect_cce}. Additionally, due to adopting lightweight auxiliary semantic decoders, the variant of ObjFormer for SCD does not necessitate too many additional parameters and computational overhead compared to the basic architecture for BCD.

\subsection{Ablation Study}

\newcommand{\reshl}[2]{
#1 \fontsize{7pt}{1em}\selectfont\color{blue}{$\!\uparrow\!$ \textbf{#2}}
}
\begin{table*}[!t]
  \renewcommand{\arraystretch}{1.3}
\caption{\centering{Ablation study for key modules in ObjFormer. Here, OGSA means object-guided self-attention, OGCA means object-guided cross-attention, MOT means multi-scale object test, and MOL refers to multi-scale object learning (training+testing).}}\label{tbl:ablation_study}  
  \centering
  \begin{tabular}{c c c c l l l l l}
  \toprule
    \hline	
\multicolumn{4}{c}{\textbf{Module}}	& \multirow{2}{*}{\textbf{Rec}}	& \multirow{2}{*}{\textbf{Pre}}	 & \multirow{2}{*}{\textbf{OA}} & \multirow{2}{*}{\textbf{F1}}  & \multirow{2}{*}{\textbf{KC}}   \\
\cline{1-4}
OGSA & OGCA & MOT & MOL & & & & &  \\
    \hline\hline
\multicolumn{4}{c}{Baseline network} & 0.8302	&0.8470	&0.8823&	0.8385&	 0.7460 \\
 $\checkmark$& & & & \reshl{0.8423}{0.0121} & \reshl{0.8607}{0.0137} 	& \reshl{0.8918}{0.0095}	& \reshl{0.8514}{0.0129}	&	\reshl{0.7663}{0.0203} \\
 $\checkmark$&  $\checkmark$& & & \reshl{0.8556}{0.0254}	& \reshl{0.8761}{0.0291}	& \reshl{0.9023}{0.0200} &	\reshl{0.8657}{0.0272} &	\reshl{0.7890}{0.0430} \\
 $\checkmark$&  $\checkmark$&  $\checkmark$& & \reshl{0.8607}{0.0305}&	\reshl{0.8814}{0.0344}&	\reshl{0.9061}{0.0238}&	\reshl{0.8709}{0.0324} & \reshl{0.7972}{0.0512} \\
 $\checkmark$&  $\checkmark$& &  $\checkmark$& \reshl{0.8728}{0.0426} & \reshl{0.8811}{0.0341} &	\reshl{0.9098}{0.0275} &	\reshl{0.8769}{0.0384} & \reshl{0.8058}{0.0598} \\

    \hline
    \bottomrule
  \end{tabular}
\end{table*}

\begin{figure}[!t]
  \centering
     \includegraphics[width=3.4in]{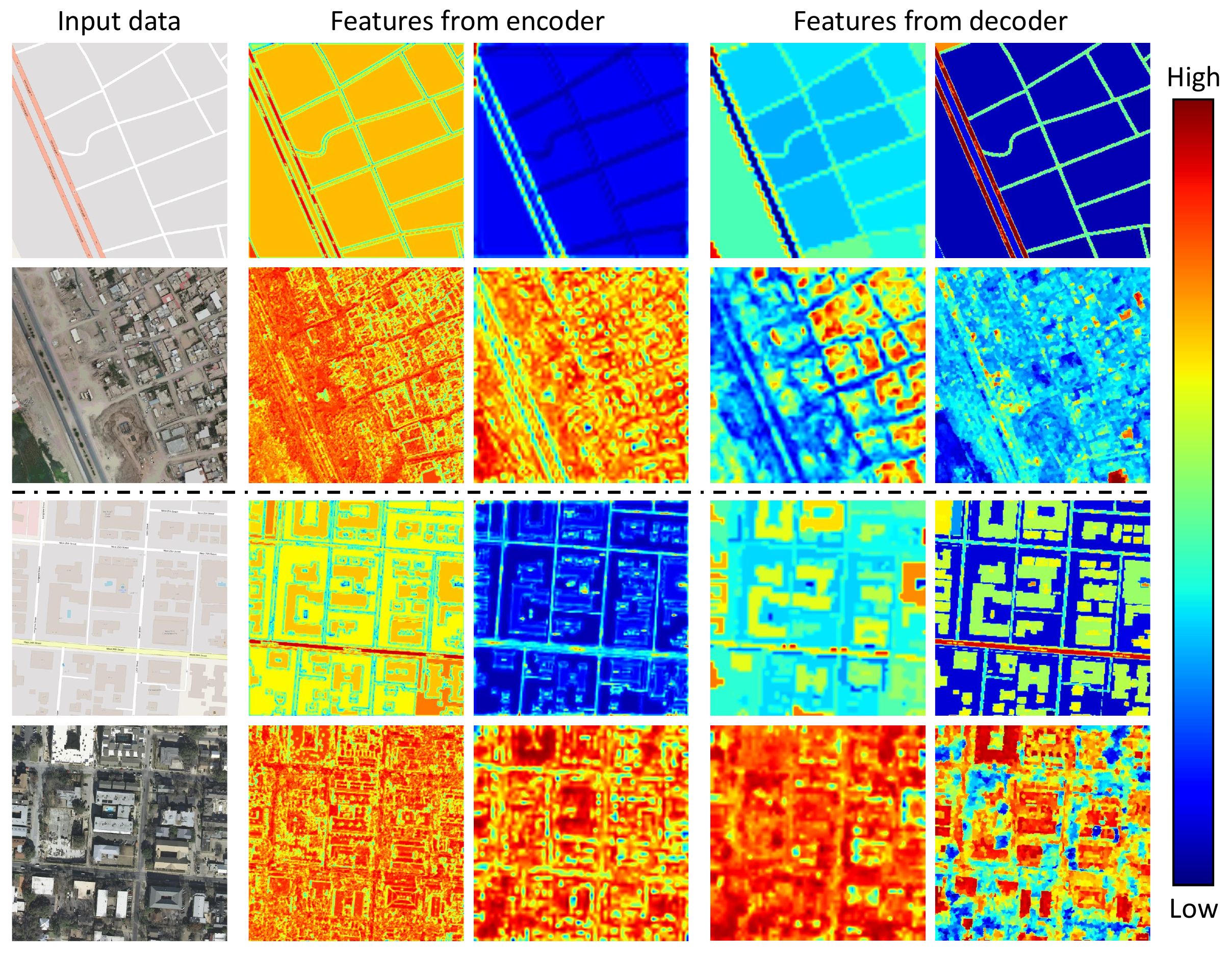}
  \caption{The visualization of deep features from different layers of ObjFormer. }
  \label{fig:visualized_feature_maps}
\end{figure}

\par ObjFormer has multiple technical improvements, including (1) integration of the self-attention module and the OBIA technique, (2) heterogeneous information fusion based on cross-attention module, (3) multi-scale object learning, and (4) learning category information from negative samples based on CCE loss. To evaluate the effectiveness of these modules, detailed ablation studies are carried out to delve into these techniques. 

\par Table \ref{tbl:ablation_study} lists the performance contribution of different key modules in ObjFormer in the BCD task. The baseline network refers to the network that does not use object-guided self-attention modules in the encoder stage and then fuses heterogeneous features in the decoder stage by using concatenation operations directly instead of using cross-attention. Since the vanilla self-attention mechanism causes a large GPU memory burden under the CD task, resulting in the network not being able to be trained with the same protocol, the baseline network employs the sequence reduction method proposed in \cite{Wang2021Pyramid} to reduce the computational and memory burden of vanilla self-attention. The training protocol and other network hyperparameters are kept the same as those of ObjFormer. We will further compare the sequence reduction method with our object-guided self-attention in Section \ref{sec:effect_OBIA}. 

\par Firstly, the proposed object-guided self-attention module can contribute to the improvement in KC of 2.03$\%$, revealing the effectiveness of our motivation to combine the OBIA method with the Transformer architecture. Then, adopting the object-guided cross-attention module to fuse the extracted hierarchical features can bring about a 2.27$\%$ improvement in KC, demonstrating the importance of well-designed fusion modules to fuse heterogeneous features under our task again. Fig. \ref{fig:visualized_feature_maps} shows the response values of the feature maps extracted from different layers of ObjFormer's encoder and decoder. Benefiting from the object-guided self-attention mechanism, the features extracted from the optical images are in terms of objects, and the edge information of the features can be well maintained. For the map data, it goes a step further, and the same instances maintain consistent feature values. In the decoder part, after continuous fusion, the same land-cover features in the feature map of optical image can keep similar response values, indicating that high-level semantic information has been well interpreted. Subsequently, the detection performance is further improved by using multi-scale object test, which brings 0.82$\%$ KC improvement. Finally, adopting multi-scale object learning in the training stage can finally make the KC value of ObjFormer to 0.8058. These results fully demonstrate the effectiveness of the individual components and modules in ObjFormer.

\subsection{Effect of Introducing OBIA for Transformer}\label{sec:effect_OBIA}
\par One of our contributions in this work is to combine the widely used OBIA technique with advanced vision Transformer architecture, thus naturally reducing the computational overhead of the self-attention mechanism while introducing low-level information to the network. Therefore, we will conduct experiments to verify the effectiveness of this point. Firstly, we will discuss the performance robustness of the proposed ObjFormer under different scales of the generated objects. Then, we will discuss how different parameter-free fusion methods $\mathcal{F\left(\cdot\right)}$ will have an impact on ObjFormer's detection performance, including taking the mean, maximum, and minimum values over each channel, respectively, and taking the mean and maximum values for each channel and summing them. Finally, we will study the specific effect of the introduction of the OBIA technique on the computational overhead of the self-attention mechanism and the total model complexity. 

\begin{figure}[!t]
  \centering
  \includegraphics[width=3.45in]{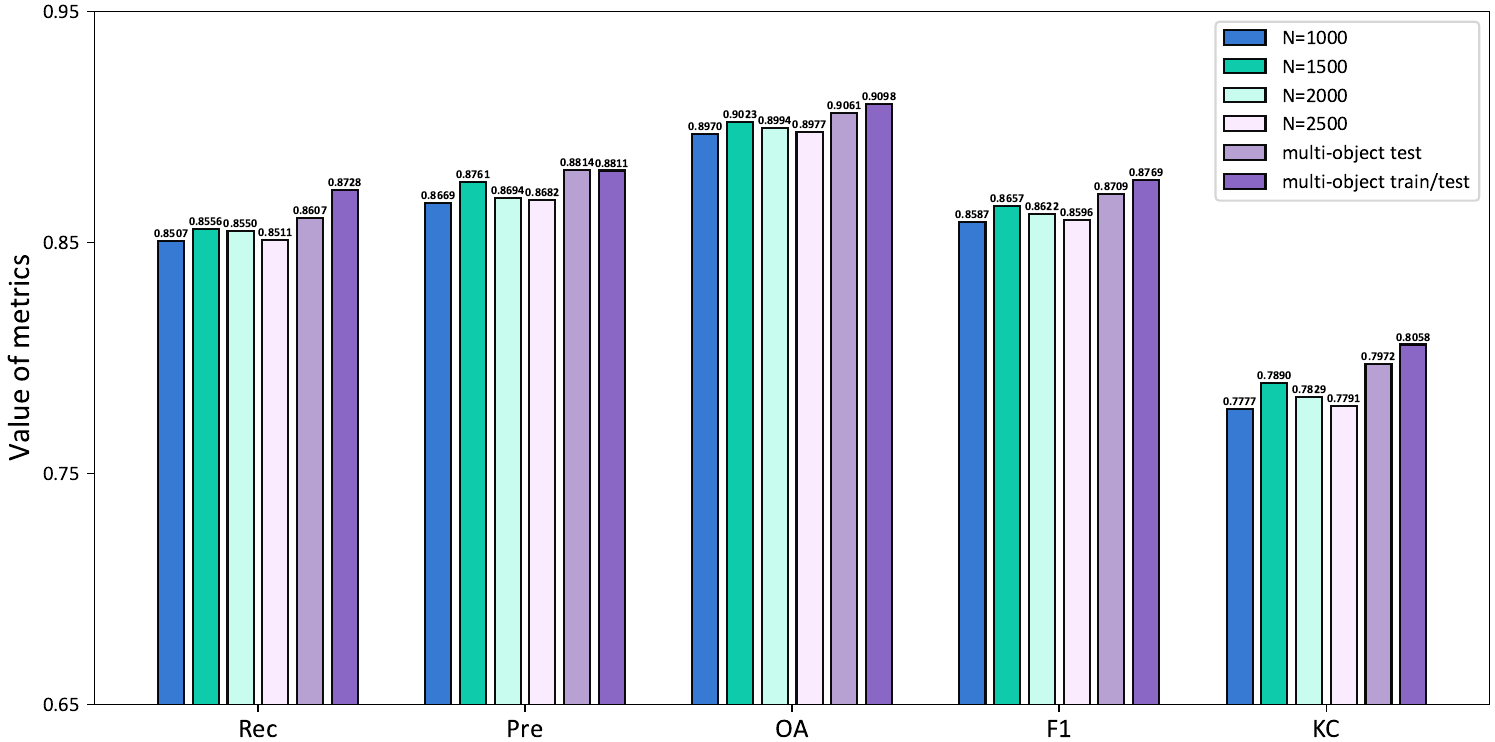}
  \caption{The relationship between the number of objects and BCD performance of ObjFormer. Here the number of objects refers to the average number of objects obtained by the SLIC algorithm for an optical imagery with an input size of 512$\times$512.}
  \label{object_scale_analysis}
\end{figure}

\subsubsection{\textbf{Different object scales}}
\par Fig. \ref{object_scale_analysis} shows the BCD performance of ObjFormer with objects of different scales. Here, we control the object's scale by adjusting the number of objects generated by the SLIC algorithm. We can see that the scale of the objects in the generated object map affects the performance of ObjFormer to some extent. In general, if the scale of generated objects is too large, they may contain several kinds of land-cover features, causing the generated deep object tokens to be not homogeneous enough, thus reducing the performance. If the object scale is too small, the computational overhead will inevitably increase. From Fig. \ref{object_scale_analysis}, the optimal object scale for ObjFormer on the OpenMapCD dataset is $N_{o}=1500$. Yet it can also be observed that the ObjFormer architecture itself has a relatively high robustness to object scales, since the KC varies no more than 1.2$\%$ in the range of $N_{o}=1000$ to $N_{o}=2500$. Finally, the determination of the optimal object scale can be avoided by using the proposed multi-scale object learning approach, and better detection results can be obtained.

\begin{figure}[!t]
  \centering
   \centering
  \subfloat[]{
     \includegraphics[width=1.64in]{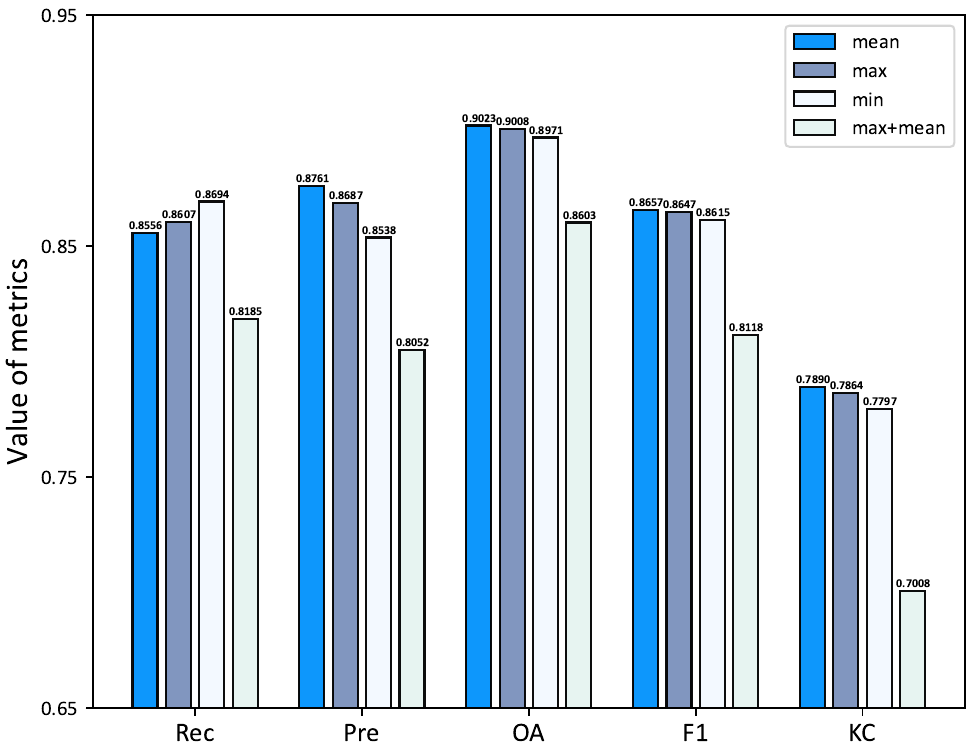}
  \label{fig_second_case}}
  \hfil
  \subfloat[]{
     \includegraphics[width=1.64in]{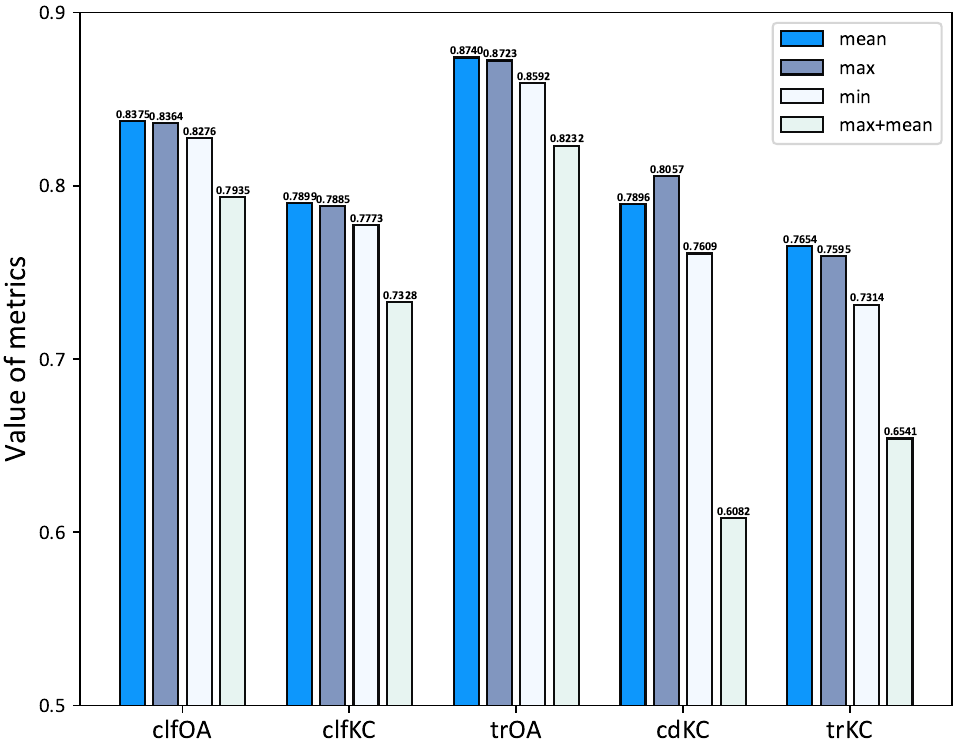}
  \label{fig_second_case}}
  \caption{The effect of different statistics for generating deep object tokens on (a) BCD and (b) SCD performance of ObjFormer. }
  \label{object_statistics_analysis}
\end{figure}

\par \subsubsection{\textbf{Different fusion methods}}
Fig. \ref{object_statistics_analysis} shows the detection performance of adopting different fusion methods $\mathcal{F}\left(\cdot\right)$ to generate deep object tokens. It can be seen that generating tokens by taking the mean value of all pixels within an object over each channel achieves the highest KC for BCD and the highest trKC for the SCD task. This is because individual objects are generally homogeneous and the pixels within the same object tend to have the same land-cover features. Thus, the tokens obtained by taking the mean values can represent the object's features well. The KC obtained by taking the maximum value of all pixels within the object in each channel is very close to the mean value. This is because the features obtained in this way are salient enough, with a similar principle to maximum pooling. However, if we use both mean and maximum statistical ways to generate deep object tokens, the performance will instead be degraded. We argue that this is because using both mean and maximum statistical ways can result in feature maps that neither retains the most salient signals as effectively as only taking maximum value nor provides representative features by only taking mean values. Interestingly, using the minimum statistic to generate deep object tokens can also achieve fair performance on binary and SCD tasks. We argue that this is because, like other Transformer architectures, ObjFormer uses GELU \cite{Hendrycks2016Gaussian} rather than ReLU as the activation function, and thus taking the minimum value on each channel of all pixels inside each object also yields meaningful features rather than all-zero features.

\newcommand{\reshld}[2]{
#1 \fontsize{7pt}{1em}\selectfont\color{blue}{$\!\downarrow\!$ \textbf{#2}}
}
\begin{table*}[!t]
  \renewcommand{\arraystretch}{1.3}
\caption{\centering{The effect of combining OBIA with the self-attention mechanism on reducing computational overhead with the input size of 512$\times$512. We were unable to train the network with vanilla self-attention under the same training protocol due to the over-large computational overhead.}}\label{tbl:OSA_com} 
  \centering
  \begin{tabular}{c c c l c}
  \toprule
    \hline	
    \textbf{Method}	& \textbf{Complexity} & \textbf{Parameters (M)}	& \textbf{MACs (G)} &  \textbf{KC}  \\
    \hline\hline
    Vanilla \cite{dosovitskiy2020image} & $O\left(HW\times HW\right)$ &	28.37	 &	223.13 & - \\
    Efficient \cite{Wang2021Pyramid} & $O\left(\frac{HW\times HW}{R^{2}}\right)$ &	34.08  &	\reshld{26.42}{196.71}  & 0.7798 \\
    Ours & $O\left(N^{obj} \times N^{obj}\right)$ &	28.37  &	\reshld{27.12}{196.01}  & 0.7890 \\
    \hline
    \bottomrule
  \end{tabular}
\end{table*}

\begin{table*}[!t]

  \renewcommand{\arraystretch}{1.3}
\caption{\centering{Effect of CCE loss for different comparison models on SCD.}}\label{tbl:cce_acc} 
  \centering
  \begin{tabular}{c l l l l l}
  \toprule
    \hline	
\textbf{Method}	& \textbf{clfOA}	& \textbf{clfKC} &		\textbf{cdKC} &	\textbf{trOA} & \textbf{trKC}  \\
\hline\hline
HRSCD-S1 w/ CCE&	\reshl{0.7474}{0.0605} &	\reshl{0.6735}{0.0729}	&  \reshl{0.6069}{0.1892} & \reshl{0.7741}{0.0652}	&	\reshl{0.6090}{0.1532} \\
HRSCD-S3 w/ CCE &	\reshl{0.7253}{0.0376}& \reshl{0.6417}{0.0418}&	 \reshl{0.6294}{0.0197}  &\reshl{0.8070}{0.0519} &	\reshl{0.6190}{0.1343} \\
HRSCD-S4 w/ CCE& 	\reshl{0.7522}{0.0713} & \reshl{0.6805}{0.0841}&		\reshl{0.6906}{0.0172} & \reshl{0.8163}{0.0593} &\reshl{0.6521}{0.1477} \\

ChangeMask w/ CCE &	\reshl{0.8332}{0.0794} & \reshl{0.7851}{0.0999} & 	\reshl{0.7785}{0.1095} &\reshl{0.8652}{0.0859} &	\reshl{0.7456}{0.2100} \\

SSCD-1 w/ CCE &	\reshl{0.8257}{0.0885}  & \reshl{0.7756}{0.1117} &  \reshl{0.7828}{0.0669}	 & \reshl{0.8581}{0.0871} &	\reshl{0.7363}{0.2158}	 \\

BiSRNet w/ CCE &	\reshl{0.8292}{0.0970}  & \reshl{0.7799}{0.1206} &  \reshl{0.7832}{0.0881} & \reshl{0.8630}{0.0887}	&	\reshl{0.7449}{0.2085} \\

TED w/ CCE &	\reshl{0.8216}{0.0790}  & \reshl{0.7708}{0.0975} & \reshl{0.7734}{0.1114} & \reshl{0.8580}{0.0858} &	\reshl{0.7392}{0.2155} \\

SCanNet w/ CCE &	\reshl{0.8330}{0.0870}  & \reshl{0.7851}{0.1101}  & \reshl{0.7839}{0.0483}	& \reshl{0.8665}{0.0884} & \reshl{0.7531}{0.2104}	 \\

\hline	
ObjFormer w/o CCE	& 0.7495 &	0.6794 & 0.7139 &	0.7856 &	0.5460 \\
ObjFormer w/ CCE	& \reshl{0.8375}{0.0880} &	\reshl{0.7899}{0.1105} & 	\reshl{0.7896}{0.0757}&\reshl{0.8740}{0.0884} &	\reshl{0.7654}{0.2194} \\
ObjFormer w/ full labels	& 0.8697 &	0.8308 & 0.8096 & 0.8973 & 0.8122 \\
    \hline
    \bottomrule
  \end{tabular}
\end{table*}

\subsubsection{\textbf{Computational overhead}}
\par Table \ref{tbl:OSA_com} compares the vanilla self-attention mechanism and a commonly used efficient self-attention mechanism \cite{Wang2021Pyramid} with our object-guided self-attention mechanism. Directly adopting the vanilla self-attention mechanism in dense prediction tasks, including CD, causes significant GPU memory burden and computational overhead. Under the same training protocol, if vanilla self-attention is employed, it will directly result in the inability to train the network for our task. The efficient self-attention is to reduce the sequence length (token numbers) before calculating the self-attention map. In this way, the complexity of self-attention can be reduced from $O\left(\mathrm{HW}\times \mathrm{HW}\right)$ to $O\left(\mathrm{HW}\times \mathrm{HW} / R^{2}\right)$, where $R$ is the reduction ratio. On the contrary, our approach naturally reduces the complexity of the self-attention mechanism from $O\left(\mathrm{HW}\times \mathrm{HW}\right)$ to $O\left(N^{obj}\times N^{obj}\right)$ by introducing the prevalent OBIA method, where $N^{obj}<<\mathrm{HW}$. MACs similar to efficient self-attention are achieved without adding any extra network parameters. Moreover, the efficient self-attention ignores the actual irregular distribution of land-cover features, thus inevitably losing much information after reducing the sequence length. In contrast, our method introduces low-level information to aid in reducing the sequence length. It is more consistent with the actual distribution of irregular land-cover features and also results in less information being lost after the sequence length is reduced, since the information contained within an object is often homogeneous, thereby achieving better detection performance.

\subsection{Effect of Converse Cross-Entropy Loss}\label{sec:effect_cce}
\par To verify the effectiveness of the proposed CCE loss for semi-supervised SCD, in addition to comparing the performance of ObjFormer trained with and without CCE loss, we will also train comparison methods with CCE loss to show its generalization. Furthermore, we will investigate the detector's performance on the changed and unchanged areas during the training process before and after adopting the CCE loss.

\subsubsection{\textbf{Performance improvements}}
\par Table \ref{tbl:cce_acc} reports the detection performance of the four comparison methods in SCD trained using the proposed CCE loss and list the improvement of each metric. It can be seen that the proposed CCE significantly improves the performance of the land-cover mapping task for the eight comparison methods, thereby contributing to the performance of SCD, with improvement rates of trKC ranging from 13.4$\%$ to 21.5$\%$. In addition, the difference in trKC between the ObjFormer trained with CCE and the ObjFormer trained with full optical image labeling is only 4.68$\%$. These results demonstrate the generalizability of CCE loss and validity of our motivation, i.e., helping models recognize what category a region is by having them learn what category it is not.

\begin{figure}[!t]
  \centering
  \subfloat[]{
    \includegraphics[width=1.67in]{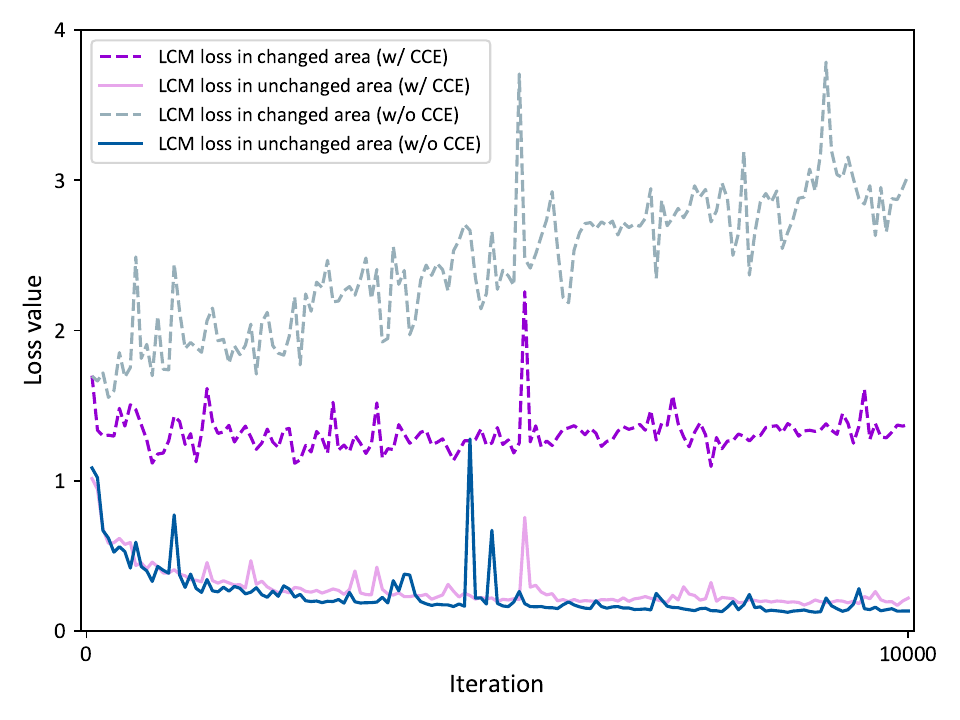}
  \label{fig_second_case}}
  \subfloat[]{
    \includegraphics[width=1.67in]{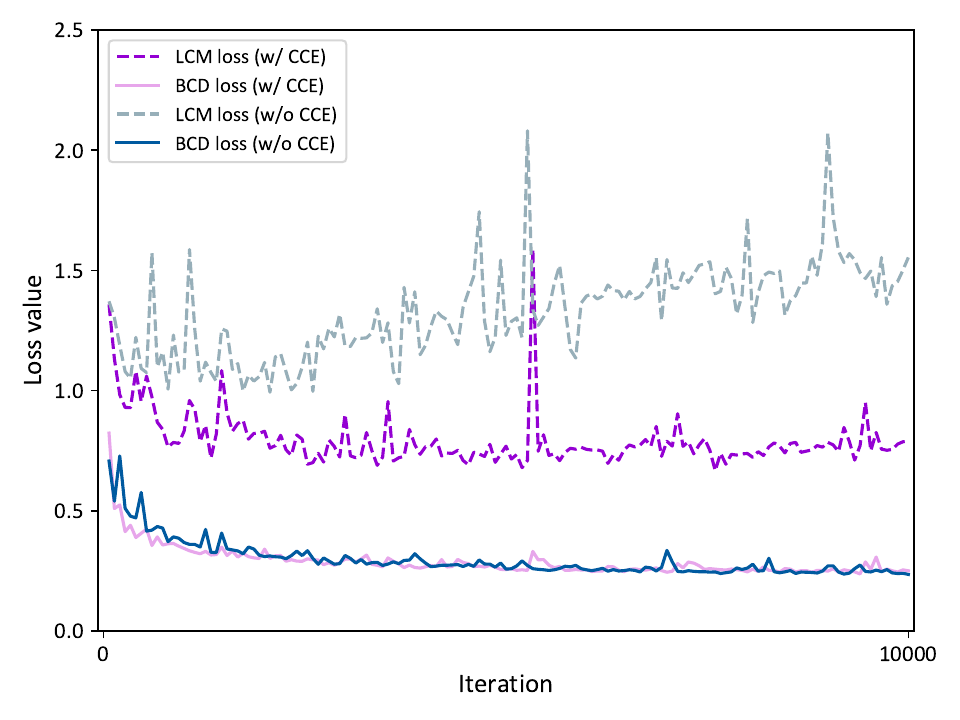}
  \label{fig_second_case}}

  \subfloat[]{
    \includegraphics[width=1.67in]{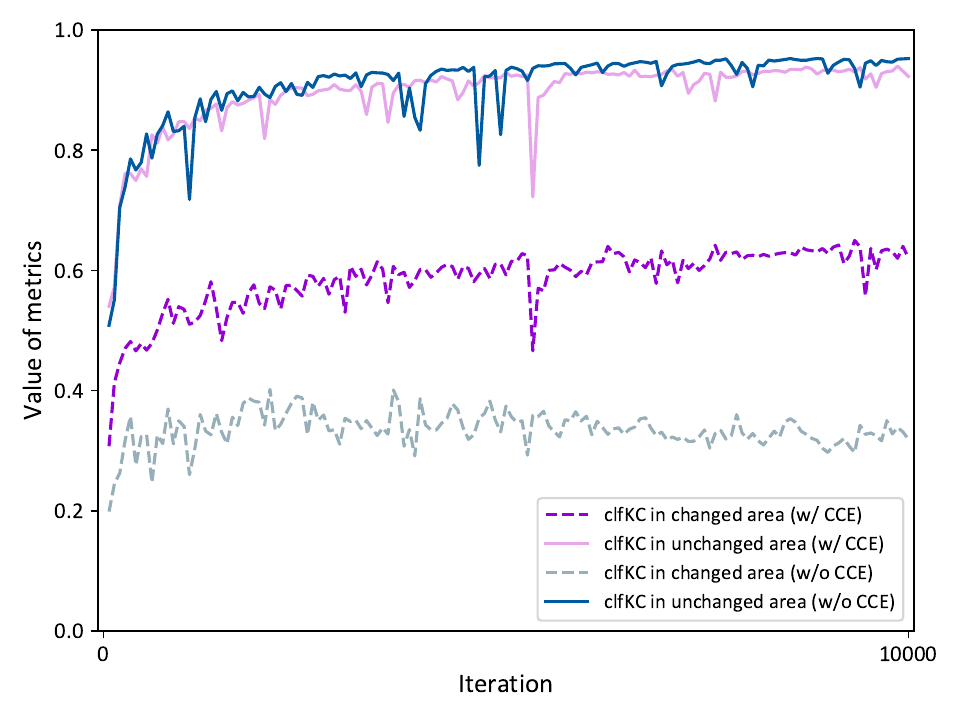}
  \label{fig_first_case}}
  \subfloat[]{
    \includegraphics[width=1.67in]{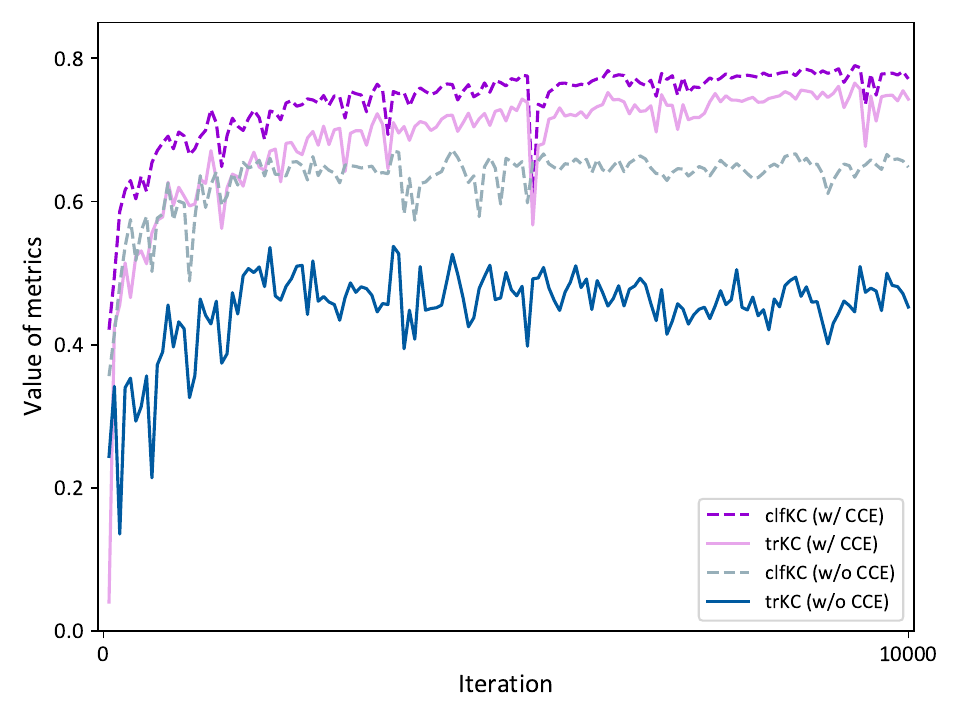}
  \label{fig_first_case}}
  \caption{Learning curves of ObjFormer trained with and without CCE loss in the SCD task. (a) Loss curves for the land-cover mapping task on the training set. (b) Loss curves for the land-cover mapping and BCD tasks on the test set. (c) Accuracy (clfKC) curves for the land-cover mapping task on the training set. (d) Accuracy curves for the land-cover mapping (clfKC) and SCD tasks (trKC) on the test set. }
  \label{fig:cce_loss_curve}
\end{figure}

\subsubsection{\textbf{Learning curves}}
\par Fig. \ref{fig:cce_loss_curve} shows the loss and accuracy curves of ObjFormer trained with and without CCE loss on the training and test sets. It can be seen that the main difference between using CCE and not using CCE in the training phase lies in the utilization of the information from the changed areas. If CCE is not used, as the training iteration increases, the loss value of the land-cover mapping task on the changed areas will gradually increase, while the corresponding clfKC gradually decreases and eventually oscillates around 0.3, as shown in Fig. \ref{fig:cce_loss_curve}-(a) and Fig. \ref{fig:cce_loss_curve}-(c). It indicates that the network's misclassification of the changed areas is more serious, leading to a final trKC of only about 0.5 on the test set, as shown in Fig. \ref{fig:cce_loss_curve}-(d). In contrast, training with CCE loss allows the network to learn useful information implicitly contained in the changed areas, i.e., that the region is impossible to be a certain category. In this way, the loss value of the land-cover mapping task is maintained in a relatively stable value, and the clfKC of the changed areas gradually increases during the training stage, as shown in Fig. \ref{fig:cce_loss_curve}-(a) and Fig. \ref{fig:cce_loss_curve}-(c). On the test set, the information obtained from the changed areas with the aid of the CCE loss, clfKC, and trKC of the network are significantly higher than that trained without the CCE loss. We argue that for some areas that are difficult to classify, information, such as knowing what category the area is impossible to be, helps to allow the network to classify it from the wrong category to the right one since the right category tends to be the one with the second highest probability in such cases \cite{Wang2022Semi}.

\subsection{Tolerance to Registration Errors}
\par Compared to pixel-based methods, object-based methods show stronger robustness to registration errors in previous studies \cite{Chen2014Assessment, Liu2021}. It is natural and interesting to wonder whether the Transformer architecture can be more tolerant to registration errors combined with OBIA technique compared to the other pixel-based methods in the context of CD on paired OSM data and optical imagery. Therefore, we carry out experiments to verify this. 

\begin{figure*}[!t]
  \centering
  \subfloat[]{
    \includegraphics[width=2.25in]{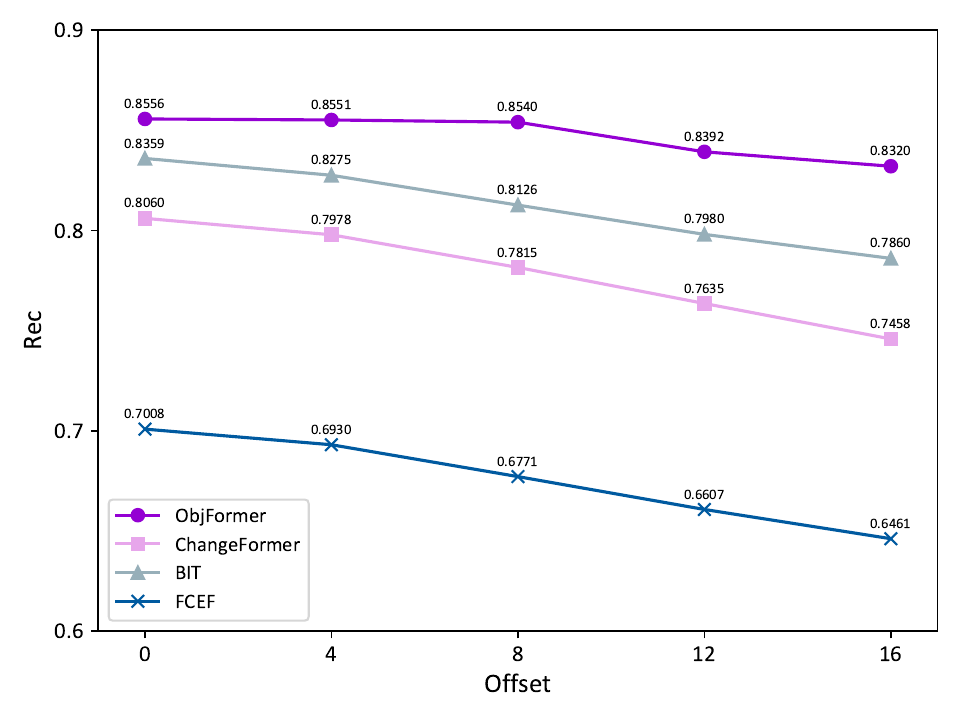}
  \label{fig_second_case}}
  \subfloat[]{
    \includegraphics[width=2.25in]{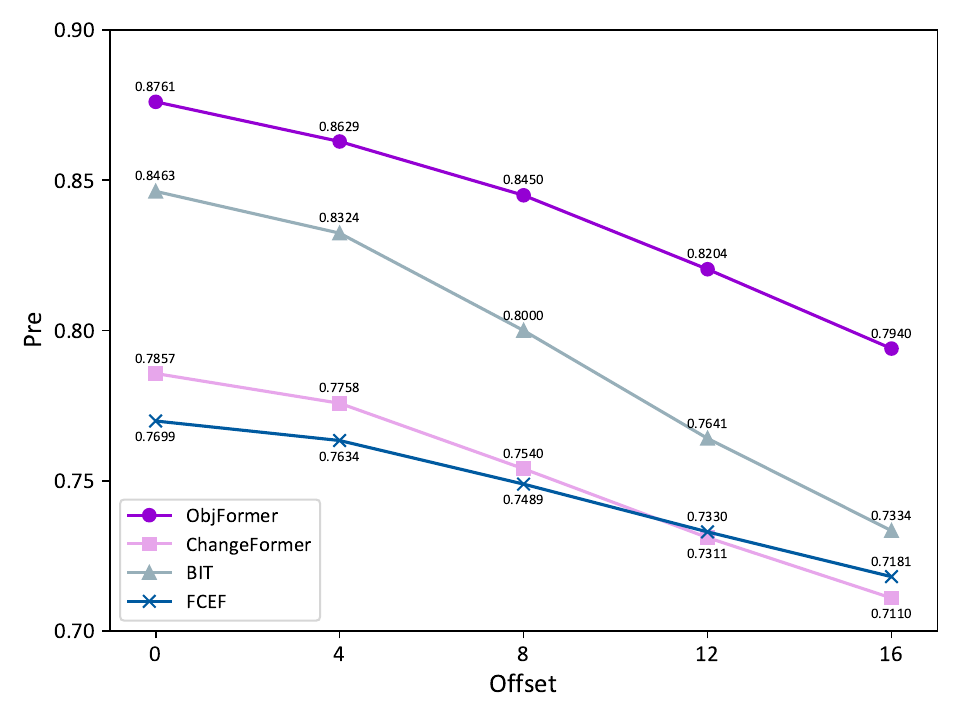}
  \label{fig_second_case}}
  \subfloat[]{
    \includegraphics[width=2.25in]{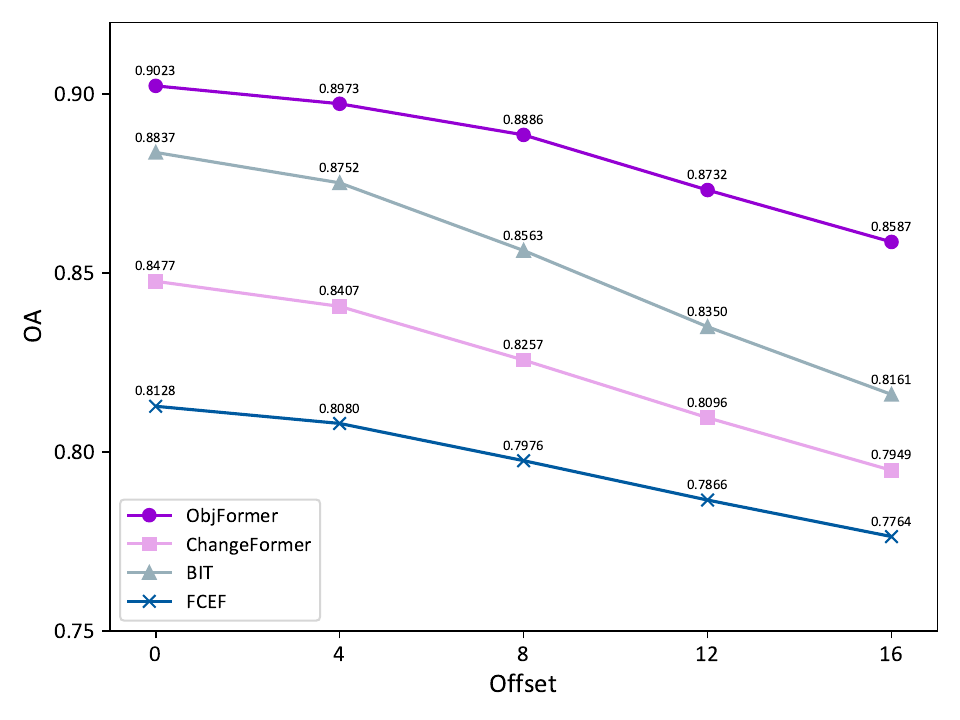}
  \label{fig_second_case}}
  
  \subfloat[]{
    \includegraphics[width=2.25in]{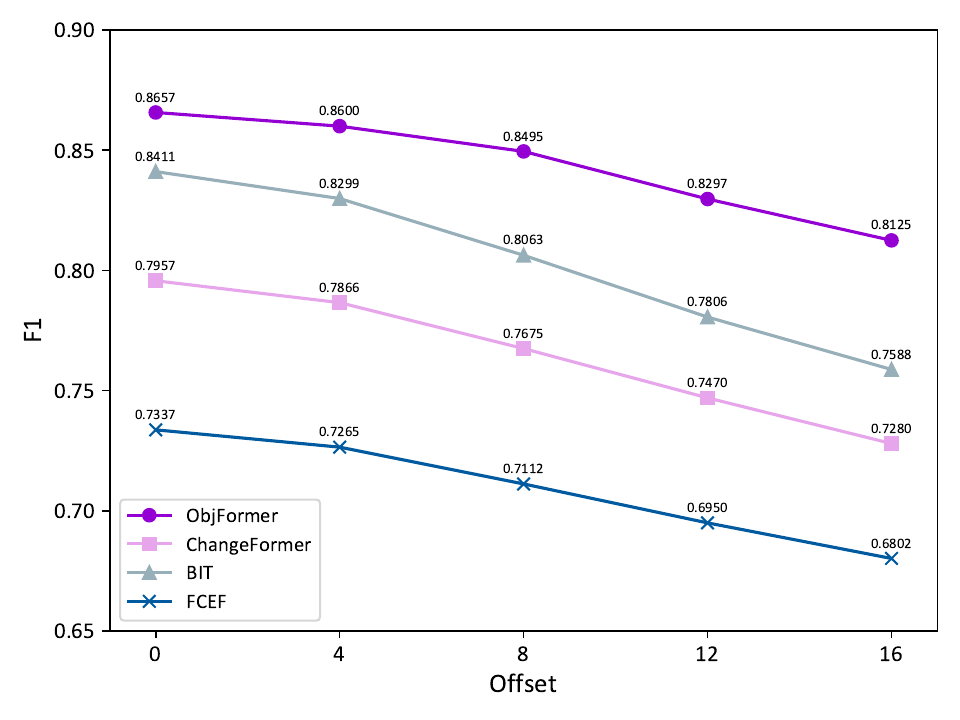}
  \label{fig_first_case}}
  \subfloat[]{
    \includegraphics[width=2.25in]{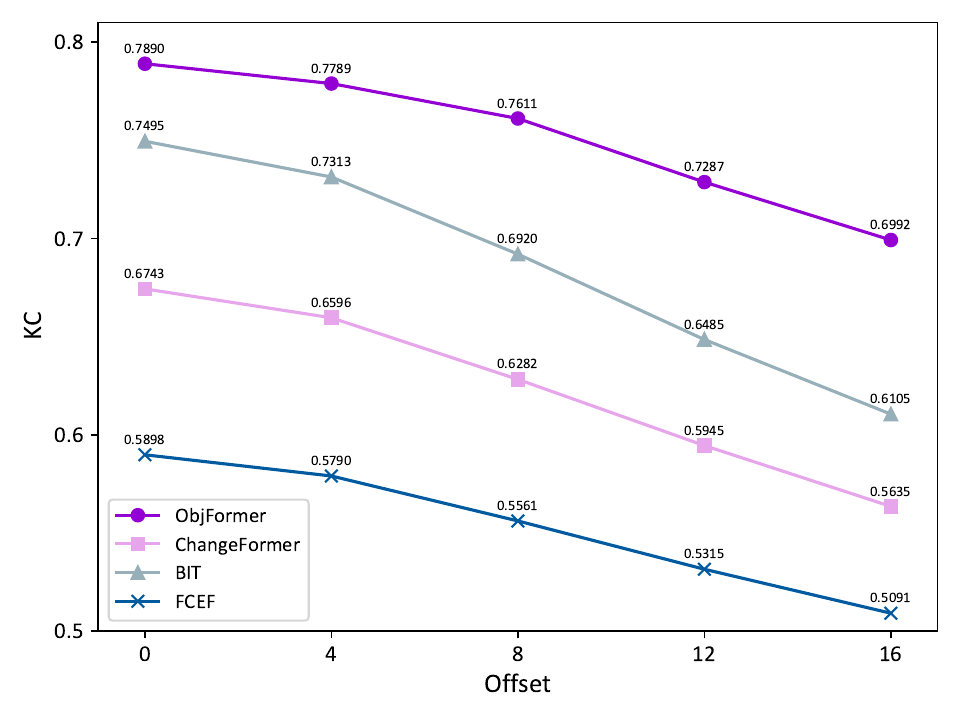}
  \label{fig_first_case}}
  \caption{Comparison of four traiend models tested on the test set with different registration errors using the five evaluation metrics, i.e., (a) recall rate, (b) precision rate, (c) overall accuracy, (d) F1 score, and (e) Kappa coefficient. }
  \label{fig:offset_test}
\end{figure*}

\subsubsection{\textbf{Testing on data with registration errors}}
\par Firstly, we want to see whether the trained detector is robust to registration errors in the test data. Therefore, we train the proposed method and comparison methods on the original training set and evaluate them on the test set with artificially introduced registration errors. We adopt a similar protocol in previous studies \cite{Townshend1992impact, Chen2014Assessment}, where we artificially generate registration errors by translating the optical image relative to the corresponding OSM data by several pixel units on the horizontal axis, vertical axis, and 45 degree angle direction, respectively. We then test the model in all three cases and take the average accuracy metrics as the final detection performance. The specific offset values in the experiment are 4, 8, 12 and 16 pixels. Fig. \ref{fig:offset_test} shows the test results of three comparison methods and our ObjFormer on the test set with different registration errors. First, the recall values of all three comparison methods undergo a relatively significant decrease as the registration error increases, whereas ObjFormer can maintain a stable recall even when the error increases to 8 pixels. For the precision rate, a noticeable decrease occurs for all methods as the registration error increases. FC-EF has the least decrease in this metric. On the comprehensive metric KC, ChangeFormer and BIT decreased by 11.08$\%$ and 13.90$\%$, respectively, while ObjFormer decreased by 8.98$\%$. It demonstrates that compared to BIT and ChangeFormer, two networks that still adopt pixels as the basic unit, the proposed ObjFormer achieves stronger robustness to testing registration errors by introducing the OBIA method. Intuitively, our object-guided self-attention method takes an object composed of homogeneous pixels as the basic unit and then takes the feature of individual pixels within the object averaged across channels as a token for self-attention modeling, which means that even if a certain registration error occurs, the features used for self-attention are still similar. Furthermore, we also note that FC-EF has slightly better robustness to test registration errors than our method, with an 8.07$\%$ decrease in KC. This reveals that the early-fusion structure could be more robust to the testing registration error than the siamese network structure. A detailed study of the robustness of different structure paradigms to registration errors can be considered in subsequent research.

\begin{figure*}[!t]
  \centering
  \subfloat[]{
    \includegraphics[width=2.25in]{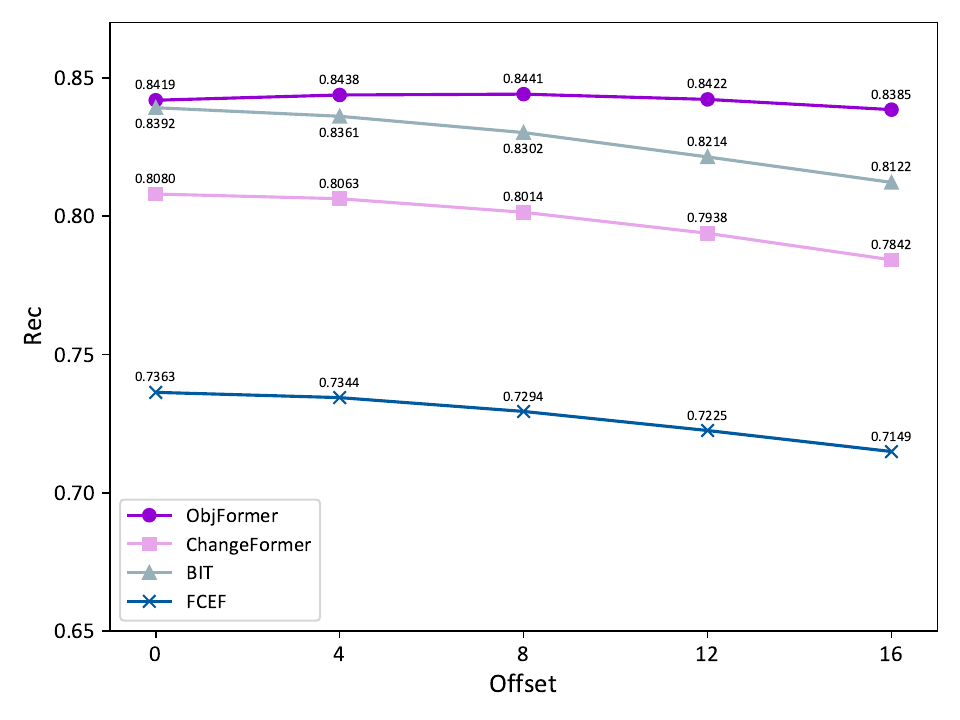}
  \label{fig_second_case}}
  \subfloat[]{
    \includegraphics[width=2.25in]{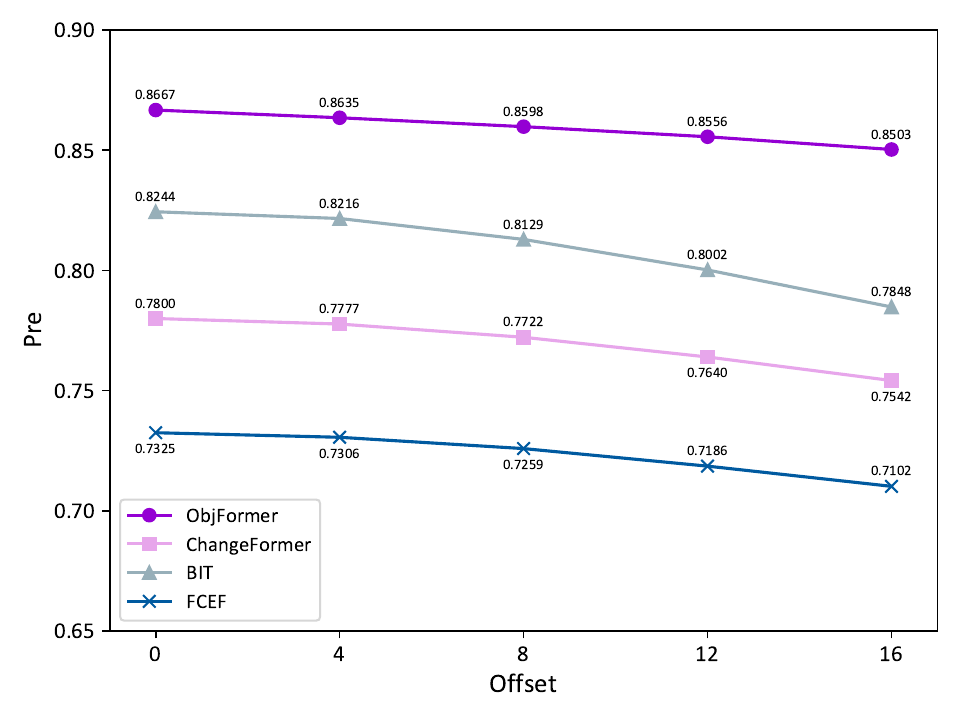}
  \label{fig_second_case}}
  \subfloat[]{
    \includegraphics[width=2.25in]{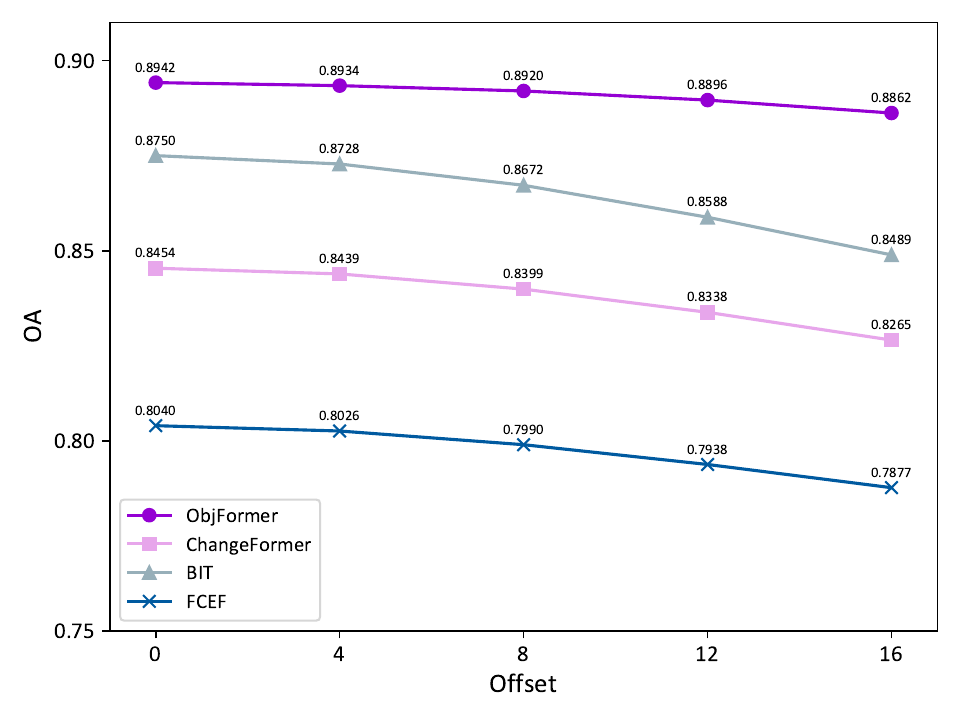}
  \label{fig_second_case}}
  
  \subfloat[]{
    \includegraphics[width=2.25in]{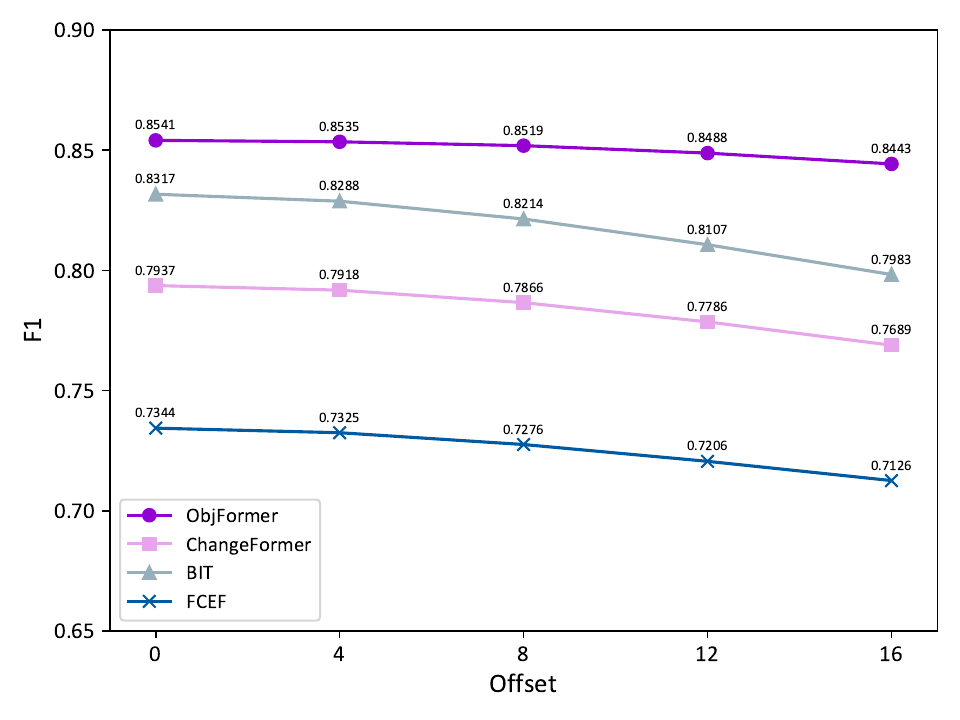}
  \label{fig_first_case}}
  \subfloat[]{
    \includegraphics[width=2.25in]{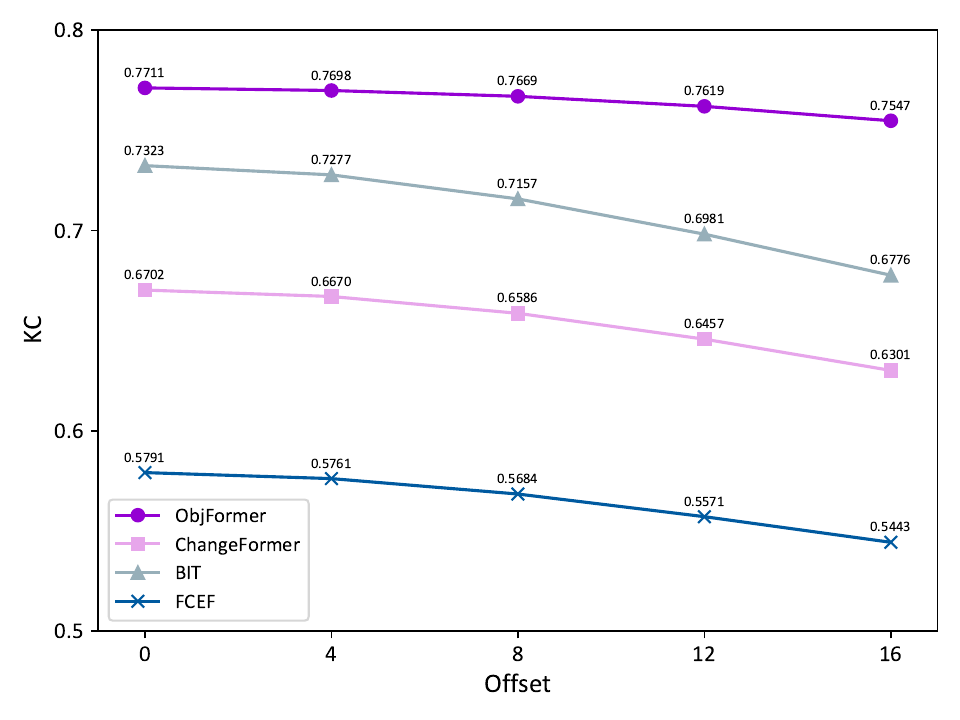}
  \label{fig_first_case}}
  \caption{Comparison of four models traiend on the training set with registration errors, tested on the test set with different registration errors using the five evaluation metrics, i.e., (a) recall rate, (b) precision rate, (c) overall accuracy, (d) F1 score, and (e) Kappa coefficient. }
  \label{fig:offset_train_test}
\end{figure*}

\subsubsection{\textbf{Training and testing on data with registration errors}}
\par Second, it is also important to investigate the effect of training data registration errors on detector performance in the context of CD on OSM data and optical images. To this end, during the training stage, we keep the OSM data fixed and randomly translate the optical images by several pixels in the horizontal axis, vertical axis, and 45-degree angle direction. Then, we evaluate detectors on the original test without registration error and test sets with simulated registration errors of varying degrees, as in the first part. Fig. \ref{fig:offset_train_test} shows the test results of the above four models trained on the data with registration errors. First, when we train the detector on the dataset with registration errors, the model's accuracy decreases on the data without registration errors (the case of offset=0), showing the negative effect of registration errors on training samples. However, detectors can obtain stronger robustness to registration error: the degree of decrease for all four models in all five metrics is lower than in Fig. \ref{fig:offset_test}. Then, we can notice that ObjFormer shows strong robustness to registration errors in this setup. In all five metrics, the degree of decrease as the registration error increases is meager. In terms of KC, ObjFormer's KC decreases by only 1.64$\%$ when the registration error increases from 0 to 16. In comparison, ChangeFormer, BIT, and FC-EF decrease by 4.01$\%$, 5.47$\%$ and 3.48$\%$, respectively. This result implies that we can use artificially created registration errors as a data augmentation means in CD tasks to improve the robustness of detectors to registration errors.

\subsection{Applications in Two Study Sites}
\par Finally, we test the proposed method on two local study sites to show its generalization ability and practical applications. To be specific, if the OSM data can represent the area at a specific time phase (i.e., comprehensively and accurately edited), then our method can be used to generate land-cover classification map of optical imagery and further analyze “from-to” change transition information for the region from OSM data and optical imagery. If the OSM data are not up-to-date or have poor editing quality, but the optical images are up-to-date, then our method can be further used to guide practitioners in updating the maps.

\begin{figure}[!t]
  \centering
  \includegraphics[width=3.49in]{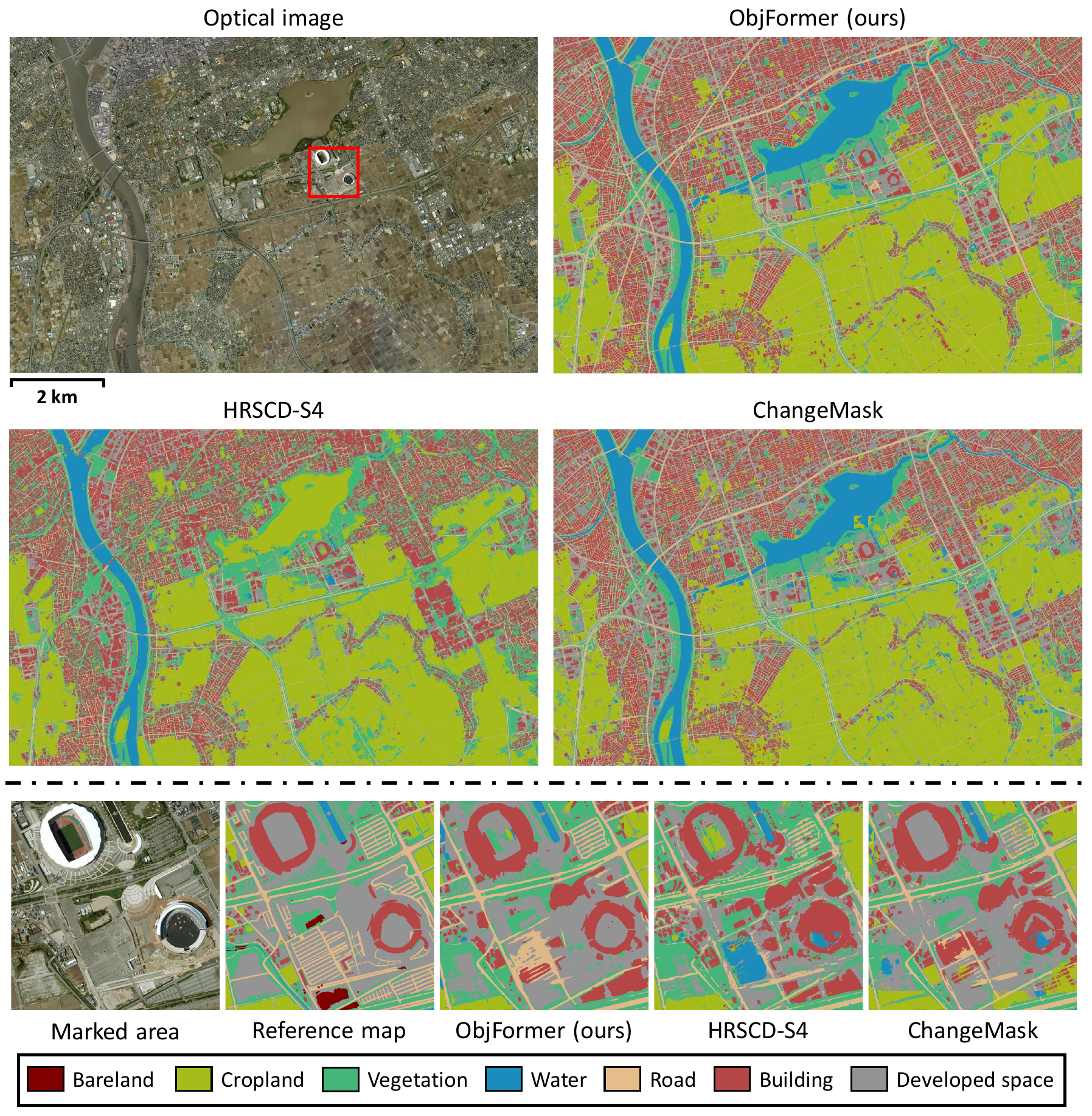}
  \caption{Land-cover classification maps on the study site I. The map products have a size of 12,036 $\times$ 18,944 pixels.}
  \label{land_coer_map_Nigata}
\end{figure}

\begin{figure}[!t]
  \centering
  \includegraphics[width=3.49in]{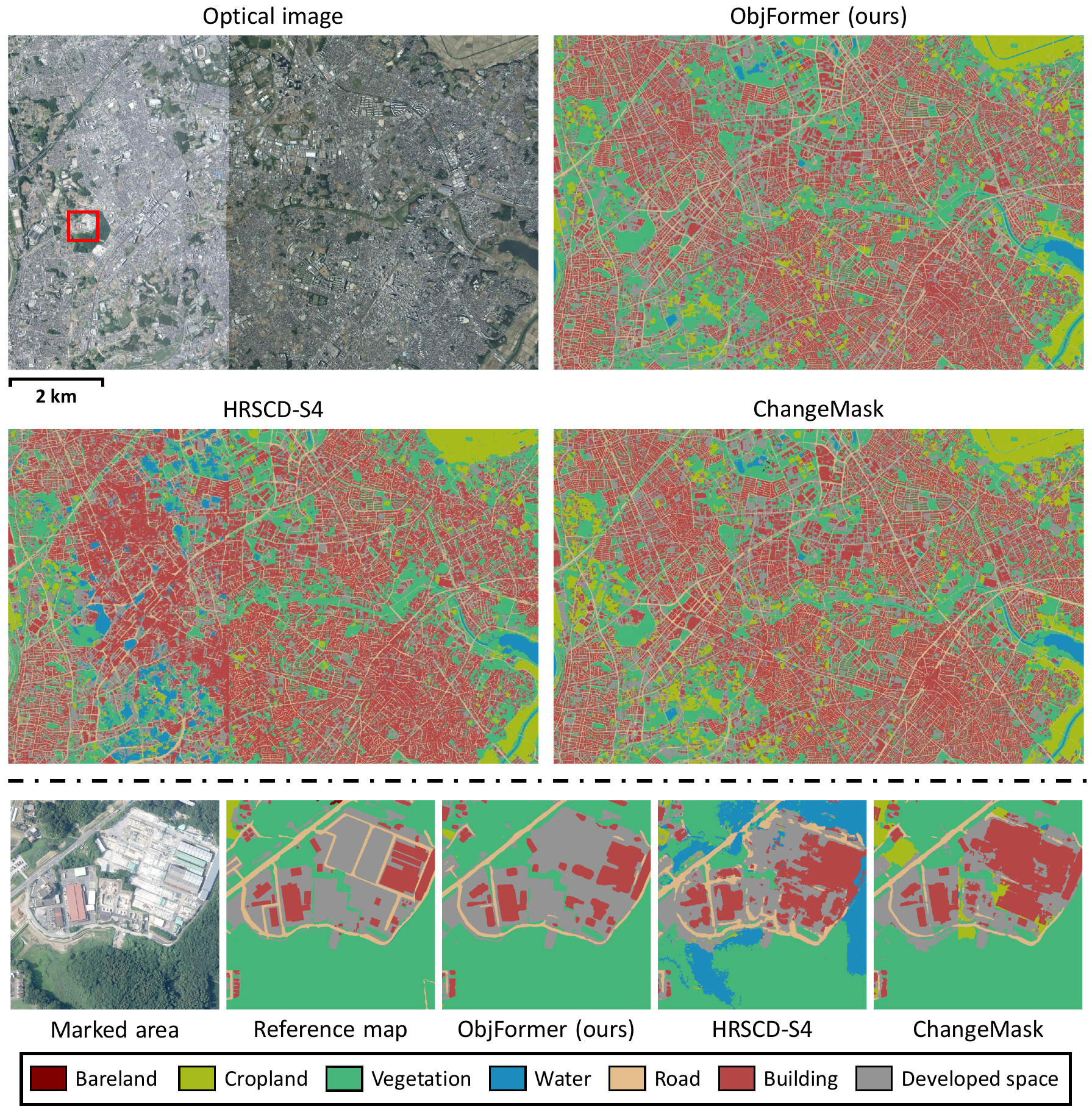}
  \caption{Land-cover classification maps on the study site II. The map products have a size of 11,776 $\times$ 18,688 pixels.}
  \label{land_cover_map_Chiba}
\end{figure}

\subsubsection{\textbf{Land-cover mapping}}
\par Fig. \ref{land_coer_map_Nigata} and Fig. \ref{land_cover_map_Chiba} display the land-cover classification maps derived through our method and two comparison methods. Notably, the optical images for the test data were captured by different sensors compared to the OpenMapCD dataset. This introduces a noticeable domain gap between the OpenMapCD dataset and the test sets, challenging the generalizability of the methodology. For example, HRSCD-S4 misclassifies lakes in the upper-middle region of study site I as croplands. In study site II, a noticeable radiometric difference exists between the left and right parts of the optical image due to their acquisition at different time phases. It can be seen that HRSCD-S4 misclassifies much of the vegetation and cropland in the left region as water bodies. Compared to the HRSCD-S4, ObjFormer and ChangeMask yield more accurate land-cover classification maps. To provide an intuitive comparison, we zoom into two specific regions in the classification results, highlighted by red boxes. For site I, we show a scene covering two stadiums in the urban area. We can see that compared to ChangeMask, the classification result obtained by our method is closer to the reference map, especially in some detail features, such as the roads in the upper left area, where our method can get the complete roads very well, while the roads obtained by ChangeMask suffer from breaks and discontinuities. For site II, we show a scene under construction. We can see that ChangeMask misclassifies some of the areas under construction as buildings, while ObjFormer can correctly classify these pixels. Moreover, other land-cover objects have clearer and more accurate boundaries in ObjFormer's classification results. These results reveal that our approach yields more precise land-cover maps for optical images across both study areas. It is crucial to note that acquiring such land-cover maps does not necessitate manually annotated pixel-wise land-cover labels of optical imagery. Instead, it only requires labels automatically generated from the OSM data according to a suitable mapping rule. This significantly diminishes the cost and enhances the usability of our framework for practical, large-scale land-cover mapping tasks.

\begin{figure*}[!t]
  \centering
  \includegraphics[width=6.4in]{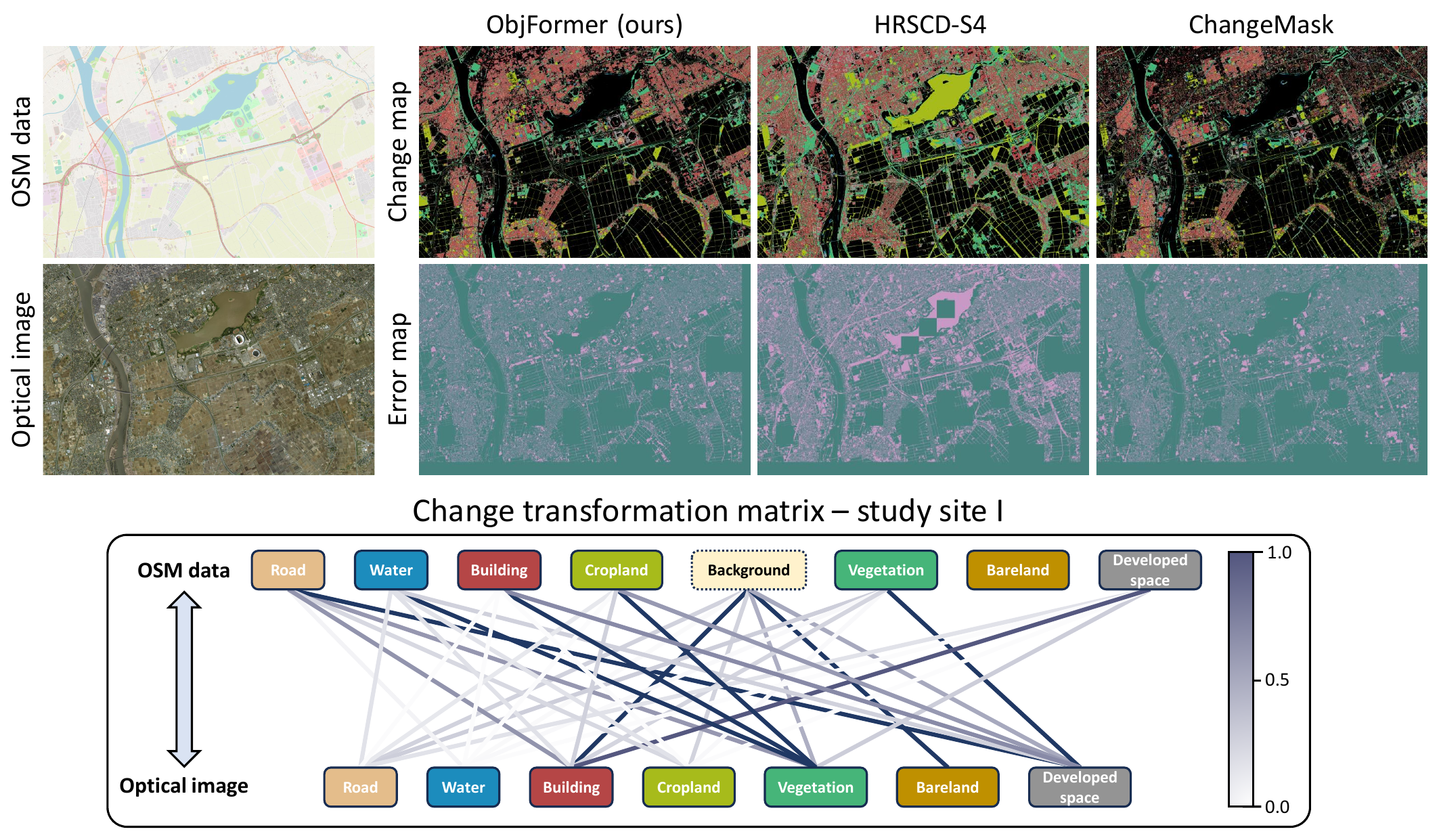}
  \caption{SCD results and change transformation matrix generated from the detection result of ObjFormer on the study site I.}
  \label{semantic_change_map_Nigata}
\end{figure*}

\begin{figure*}[!t]
  \centering
  \includegraphics[width=6.4in]{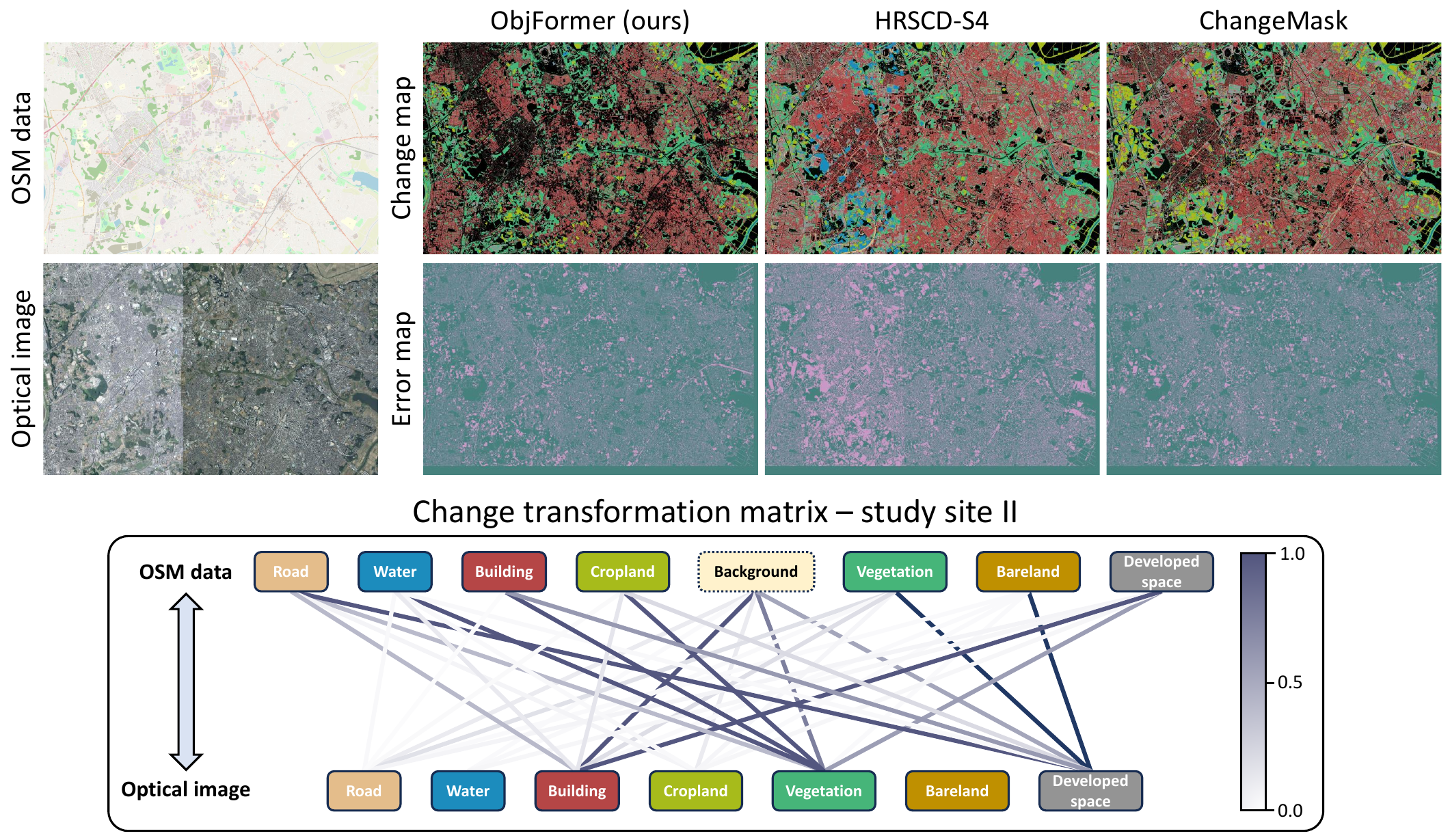}
  \caption{SCD results and change transformation matrix generated from the detection result of ObjFormer on the study site II. }
  \label{semantic_change_map_Chiba}
\end{figure*}

\subsubsection{\textbf{Semantic change analysis}}
\par Fig. \ref{semantic_change_map_Nigata} and Fig. \ref{semantic_change_map_Chiba} show further SCD results and corresponding error maps of ObjFormer and two comparison methods. Due to the high number of un-updated and unedited areas in the OSM maps of the two selected study areas. The percentage of base maps for the two regions is 28$\%$ and 53$\%$, respectively. Therefore, we can see that a vast number of pixels are changed pixels (especially in study area II). It can be seen that our method produces SCD results with fewer error pixels compared to the two compared methods, further demonstrating the generalization of our method. In addition, to visualize the change transformation information of the two study sites more intuitively, we plot the corresponding change transformation matrices based on ObjFormer's detection results. These visualizations provide insights into land-cover transitions between two time phases within the study sites. Additionally, the background category (i.e., the basemap area that did not get edited) is specifically listed in the change transformation matrix. In both study sites, a substantial portion of the basemap areas is transferred to the building areas, aligning with the observations\footnote{https://qiita.com/kouki-T/items/9b0e72710f3e8ca3dc4c} that numerous buildings in OSM in Japan are not effectively edited. Consequently, our method can provide reliable results for updating buildings in the corresponding historical map data.

\begin{figure}[!t]
  \centering
  \includegraphics[width=3.5in]{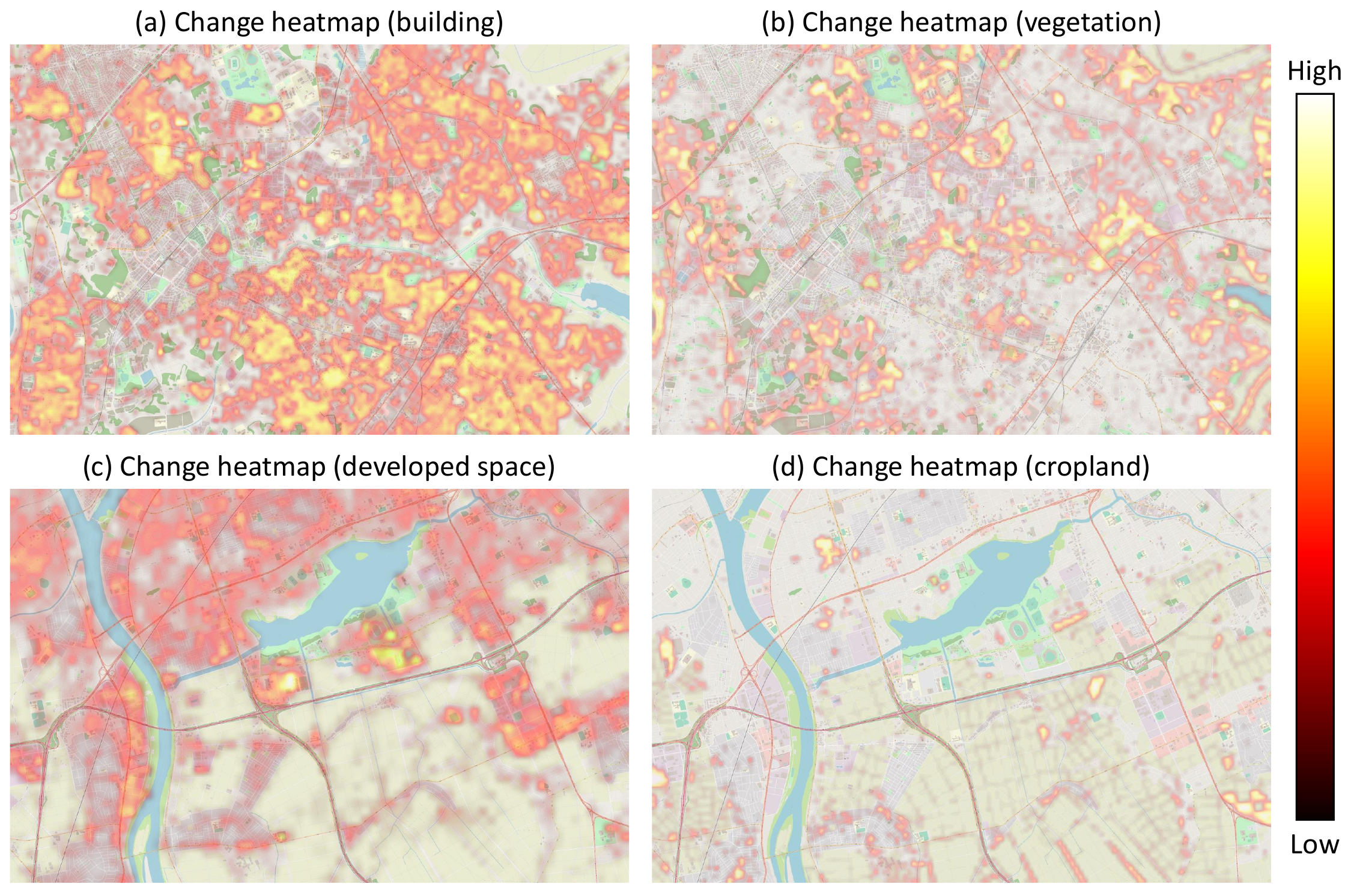}
  \caption{Heatmap products inferred from detection results, which can be used to help update map efficiently.}
  \label{heatmap_product}
\end{figure}

\subsubsection{\textbf{Map updates}}
\par The above SCD results imply that our method can be used as an effective means to aid GIS updating. The quality of OSM data, influenced by its crowdsourcing mechanism, exhibits variability. While some regions are updated proficiently, others lag due to poor updates. In Japan, for instance, despite a growing number of participants, the number of regions receiving timely updates remains limited, just like the two study sites. Our framework can furnish pertinent products for OSM participants, aiding them in swiftly updating maps. Fig. \ref{heatmap_product} illustrates the heat maps for updating the corresponding land-cover categories in the two study areas, assuming that the optical imagery is more current. By generating the heat maps in Fig. \ref{heatmap_product}, we can assist participants in quickly identifying the candidate areas most in need of updating on a large scale, as opposed to locating them through inefficient visual interpretation, thereby enhancing the efficiency of updating geographic information systems.

\begin{table*}[!t]
  \renewcommand{\arraystretch}{1.3}
\caption{\centering{Comparison of SCD performance obtained by ObjFormer and the other two models trained on the OpenMapCD dataset tested on the study sites I and II.}}\label{tbl:Niigata_acc} 
  \centering
  \begin{tabular}{c c c c c c c c c}
  \toprule
    \hline	
\multirow{2}{*}{\textbf{Study site}} & \multirow{2}{*}{\textbf{Method}} & \multirow{2}{*}{\textbf{clfOA}}	& \multirow{2}{*}{\textbf{clfKC}} &	\multirow{2}{*}{\textbf{cdKC}} &	\multirow{2}{*}{\textbf{trOA}} &	\multirow{2}{*}{\textbf{trKC}}  & \multicolumn{2}{l}{\textbf{Inference time (s)}} \\
\cline{8-9}
& & & & & & & CPU & GPU \\
\hline\hline
\multirow{3}{*}{Site I} & HRSCD-S4 &	0.6274 & 0.5394 &  0.6735 & 0.6273 &  0.5294 & 401.5 & 56.62	\\
 & ChangeMask & 0.7655	 & 0.7116 &  0.8137 & 0.7655 & 0.6886 & 754.2 & 44.24	\\
 & ObjFormer &	\textbf{0.7827} & \textbf{0.7330} & \textbf{0.8186} & \textbf{0.7728} & \textbf{0.6980} & 1058.1 & 48.81	\\
    \hline
\multirow{3}{*}{Site II} & HRSCD-S4 &	0.5861 & 0.4643 & 0.6564& 0.5861 & 0.5122 & 367.7 & 56.08 \\
& ChangeMask & 0.7192   & 0.6365  &\textbf{0.8045}  & 0.7192 & 0.6636& 701.2 & 40.41	\\
& ObjFormer &	\textbf{0.7428} & \textbf{0.6636} & 0.8030 &  \textbf{0.7379} &  \textbf{0.6847} & 956.0 & 53.34	\\
    \hline
    \bottomrule
  \end{tabular}
\end{table*}

\subsubsection{\textbf{Accuracy assessment}}
\par Table \ref{tbl:Niigata_acc} reports the specific SCD performance of our approach and two comparison methods on the two study sites. Our method achieves the highest accuracy on the land-cover mapping task and the SCD task, with clfKC of 0.7330 and trKC of 0.6980 and 0.6847 on the study site I and II, respectively, thoroughly showcasing the generalization ability of our method. Furthermore, concerning the runtime, our method exhibits a similar inference speed as the other two compared methods on the GPU. However, on the CPU, due to the Transformer architecture, our method is more time-consuming compared to the two CNN-based comparison methods.

\section{Limitations $\&$ Future Study}\label{sec:6}
\par First, the OBIA method currently generates a predefined number of objects for all optical high-resolution images, a strategy that may not optimally cater to the diverse complexity present in different scenes. For instance, urban commercial districts exhibit a more intricate landscape than suburban farmlands. For the former, we should generate higher fine-grained object maps, while for the latter, we can generate coarse-grained object maps to consider both the efficiency and performance of the network. Recognizing this, it is worthwhile to study adaptively tailoring the granularity of the object maps to the specific characteristics of the scene. 

\par For the deep features extracted at different levels, we directly sampled the corresponding object/instance map to the size of the feature map using the nearest neighbor algorithm. This approach, while functional, may not always ensure a precise alignment between the scaled object map and the actual object distribution at the respective resolution. Consequently, devising a more sophisticated sampling strategy and aligning the object map with the feature map stand as an interesting topic for subsequent research. In addition, reducing the spatial dimensions of feature maps by using deformable convolution \cite{Dai_2017_ICCV} instead of stride convolution can be more effective in retaining information because the former has a more flexible receptive field.

\par We rasterized the OSM data in this paper, treating them as images with modalities in a domain different from that of optical images. In the future, we want to take a step further in multimodal information processing by combining natural language processing and computer vision techniques to directly take the raw vector data of OSM as input. Also, our current products can only locate areas on the map that are in need of updating. A more attractive future study would be updating the content of the map directly from the latest remote sensing imagery.

\par Finally, the constructed dataset covers 40 regions on six continents. However, 24 (60$\%$) of these regions are in Europe. In the future, we will continue to expand the data amount in the dataset to include more study sites from different continents to further enrich the geographic diversity of the dataset, making it a more effective and practical benchmark. In addition, the current dataset only considers general land-cover change events. In the future, we intend to include natural and man-made disaster events in the consideration of the dataset.

\section{Conclusion}\label{sec:7}
\par This paper is concerned with supervised BCD and semi-supervised SCD using paired OSM data and optical high-resolution imagery, pioneering in the field of CD. An object-guided Transformer architecture, called ObjFormer, is proposed for these tasks, which seamlessly integrates the established OBIA technique with the cutting-edge vision Transformer architecture. Such a combination considerably reduces the computational overhead and memory burden in the original vision Transformer architecture caused by the self-attention mechanism. Specifically, the proposed architecture consists of a hierarchical pseudo-siamese encoder and a heterogeneous information fusion decoder. The encoder can extract multi-level representative features for specific data forms, and the decoder can recover the land-cover changes from the extracted heterogeneous features. For semi-supervised SCD, two auxiliary semantic decoders are added for the land-cover mapping task. To enhance the performance, we further propose a converse cross-entropy loss function for the effective utilization of negative samples. 

\par On the constructed OpenMapCD dataset, we carried out detailed experiments. ObjFormer achieved 0.8059 KC on the BCD task and 0.7651 trKC on the SCD task, outperforming current SOTA CD models. Experiments for object-guided self-attention show that introducing OBIA into the visual transformer can greatly reduce the computational overhead, reducing 87.84$\%$ MACs of vallina self-attention while guaranteeing the detection performance. Apart from that, the introduction of OBIA gives ObjFormer a stronger robustness to registration errors. Furthermore, the proposed CCE can effectively enhance the performance of semi-supervised SCD for different models. Finally, real-world applications were demonstrated through case studies in Niigata and Kashiwa, Japan, underscoring the method's versatility and its promising potential in practical applications such as land-cover mapping, semantic change analysis, and map updating.

\ifCLASSOPTIONcaptionsoff
  \newpage
\fi

\bibliographystyle{IEEEtran}
\bibliography{ObjFormer.bib}

\end{document}